\pdfoutput=1
\documentclass[11pt]{article}

\usepackage[final]{acl}

\usepackage{times}
\usepackage{latexsym}

\usepackage[T1]{fontenc}
\usepackage[utf8]{inputenc}

\usepackage{microtype}

\usepackage{inconsolata}

\usepackage{graphicx}
\usepackage{blindtext}
\usepackage{latexsym}
\usepackage{enumerate}
\usepackage{multirow}
\usepackage{multicol}
\usepackage{booktabs}
\usepackage{tabularray}
\usepackage{comment}
\usepackage{amsmath}
\usepackage{adjustbox}
\usepackage{subcaption}

\newcommand{\bo}[1]{\textcolor{red}{#1}}

\usepackage{xcolor,colortbl}
\newcolumntype{a}{>{\columncolor{pink}}c}

\title{Understanding Position Bias Effects on Fairness in Social Multi-Document Summarization}

 \author{Olubusayo Olabisi \and Ameeta Agrawal\smallskip\\
         Portland State University\smallskip\\
\texttt{\{oolabisi,ameeta\}@pdx.edu}\\}

\begin{document}
\maketitle
\begin{abstract}
Text summarization models have typically focused on optimizing aspects of quality such as fluency, relevance, and coherence, particularly in the context of news articles. However, summarization models are increasingly being used to summarize diverse sources of text, such as social media data, that encompass a wide demographic user base. It is thus crucial to assess not only the quality of the generated summaries, but also the extent to which they can fairly represent the opinions of diverse social groups. Position bias, a long-known issue in news summarization, has received limited attention in the context of social multi-document summarization. We deeply investigate this phenomenon by analyzing the effect of group ordering in input documents when summarizing tweets from three distinct linguistic communities: {\em African-American} English, {\em Hispanic-aligned} Language, and {\em White-aligned} Language. Our empirical analysis shows that although the textual quality of the summaries remains consistent regardless of the input document order, in terms of fairness, the results vary significantly depending on how the dialect groups are presented in the input data. Our  results suggest that position bias  manifests differently in social multi-document summarization, severely impacting the fairness of summarization models. 
\end{abstract}

\section{Introduction}

As the use of natural language processing models gets more prevalent in various industries, academic and social settings, it is imperative that we assess not only the quality of these models but also their fairness when exposed to data originating from diverse social groups \cite{czarnowska2021quantifying}. 
Text summarization models, in particular, facilitate the processing of large collections of a wide variety of text data by distilling documents into short, concise, and informative summaries while preserving the most relevant points from the source document \cite{nallapati2017summarunner, zhang2018neural, liu2019text}. Multi-document summarization (MDS) is the task of generating a coherent summary from a set of input documents, usually centered around a topic, as opposed to single document summarization (SDS) which takes one document as input. The input in MDS consists of multiple documents, that may have been written by distinct users, varying in linguistic diversity, styles, or dialects. 

MDS can be of type {\em extractive}, where the models extract the salient points directly from the source document to form the summary, or of type {\em abstractive} where the models generate summaries by rewriting salient information using novel words or phrases. In both cases, the resulting summary should be of good quality in terms of informativeness, coherence and relevance to the source document. At the same time, a good summary should be \emph{unbiased} and should reflect the diversity of thoughts and perspectives present in the source documents.

The notion of fairness describes equal or fair treatment without favoritism or discrimination. However, plenty of evidence suggests intrinsic societal biases in language models \cite{bolukbasi2016man,bommasani2021opportunities,deas2023evaluation}. More specific to the task of summarization, fairness is measured by the ability of algorithms to capture the peculiarity in all represented groups \cite{shandilya2018fairness,dash2019summarizing,keswani2021dialect,olabisi2022analyzing,ladhak2023pre}.

\begin{figure}[t]
  \centering
  \includegraphics[width=0.55\textwidth,trim={5cm 0 0 0},clip]{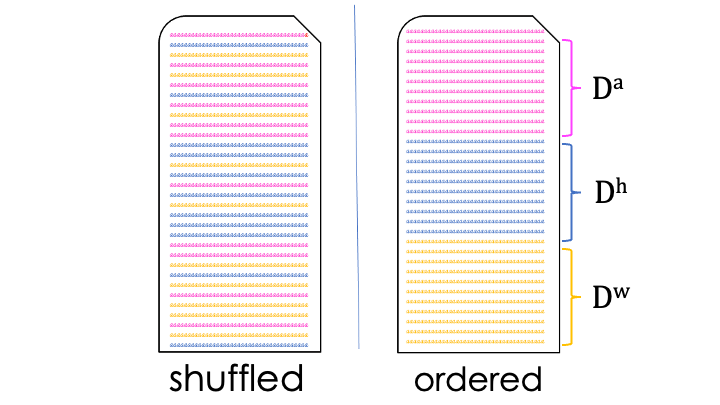}
  \caption {Illustration showing \texttt{shuffled} vs. \texttt{ordered} input for multi-document summarization consisting of documents from three diverse groups ($\mathcal{D}^a$, $\mathcal{D}^h$, $\mathcal{D}^w$) as indicated by the three colors. The \texttt{ordered} input is denoted as $\mathcal{O}^a$ when $\mathcal{D}^a$ documents appear first in the input.}
   \label{fig:Ordered.png}
\end{figure}

Conventionally, the documents in MDS  are simply concatenated into one large collection of text as the input for the model. Prior research supports the existence of position bias, or lead bias, where the models rely excessively on the position of the sentences in the input rather than their semantic information \cite{lin1997identifying, hong2014improving, wang2019exploring}. This is a particularly common phenomenon in news summarization, where early parts of an article often contain the most salient information. While many algorithms exploit this fact in summary generation, it can have a detrimental effect when important information is spread throughout the input.

In non-news domains, weak or no position bias has been observed \cite{kedzie-etal-2018-content,kim-etal-2019-abstractive}. Regardless of whether position bias is noted or not, previous investigations have quantified the {\em effects} of position bias mostly in terms of standard summarization metrics (e.g., ROUGE) which focus on the textual quality of the summary \cite{sotudeh-etal-2022-mentsum,scire-etal-2023-echoes}. In this work, we investigate the effects of position bias on the fairness of the generated summaries. 

Specifically, we ask two questions: {\em (i)} Do the system summaries show any position bias when we vary the order of the input documents? {\em (ii)} What is the impact of position bias on the fairness of the system summaries?

For our experiments we use DivSumm, a summarization dataset of linguistically diverse communities representing three dialect groups  \cite{olabisi2022analyzing}. We explore the effects of position bias in the outputs of seven abstractive summarization models (and three extractive models) and under two investigation setups: \texttt{shuffled} (when the data is presented as randomly shuffled) and \texttt{ordered} (when the input documents are grouped according to their dialects). Figure~\ref{fig:Ordered.png} presents a schematic overview. The generated summaries are evaluated in terms of fairness, as well as metrics related to the textual quality. 

The contributions of our work are as follows:
\begin{itemize}
    \item We comprehensively investigate the phenomenon of position bias in the context of social multi-document summarization;
    \item We explore ten different summarization models, both abstractive and extractive;
    \item We contextualize and quantify the impact of position bias in terms of fairness and textual quality of generated summaries.
\end{itemize}

\section{Related Work}

In this section we present some notable prior research in two relevant areas. First, we discuss position bias in summarization, followed by works studying fairness in summarization.

\medskip

\noindent \textbf{Position Bias in Summarization} \quad Position bias can manifest in MDS scenarios just as it does in SDS scenarios because in MDS, the documents are typically concatenated into one long input and treated very much like a `single' document. Several works have studied the substantial position bias (also known as lead bias), especially in the context of news summarization  where the datasets and models prioritize selecting sentences from the beginning of an article \cite{lin1997identifying, hong2014improving, wang2019exploring}. Often the lead bias is so strong that the simple lead-$k$ baseline or using the first $k$ sentences of a news article to generate the summary can score higher than many other models \cite{see-etal-2017-get}. %Shuffling the data usually helps  \cite{xu-etal-2020-unsupervised}. 
While some have suggested approaches for mitigating or countering lead bias \cite{grenander2019countering,xing-etal-2021-demoting,gong2022improving,zhang-etal-2022-attention}, others have leveraged lead bias \cite{yang-etal-2020-ted,zhu2020make,padmakumar-he-2021-unsupervised}.

Interestingly, although position bias dominates the learning signal for news summarization or similar domains, it is less apparent in other domains where most non-news datasets show {weak or no position bias}  \cite{kedzie-etal-2018-content,jung2019earlier,kim-etal-2019-abstractive,sharma-etal-2019-bigpatent,sotudeh-etal-2022-mentsum,scire-etal-2023-echoes}. Notably, none of these studies consider datasets where data originates from diverse social groups, which is the focus of our work.

Moreover, prior research studying the effect of position bias has quantified its impact exclusively in terms of textual quality, typically measured in terms of summarization metrics such as ROUGE, and others. To our knowledge, ours is the first work quantifying the impact of position bias in multidocument summarization in terms of fairness where data originates from diverse social groups.

\medskip

\noindent \textbf{Fairness in Summarization} \quad A significant amount of work has been done toward improving the textual quality of summaries but not so much in terms of enhancing the  fairness of summaries, particularly in the context of diverse groups. Prior text summarization work has proposed fairness-preserving algorithms \cite{shandilya2018fairness,dash2019summarizing}, bias mitigation models \cite{keswani2021dialect} and fairness interventions for extractive and abstractive summarization \cite{olabisi2022analyzing}. Furthermore, \citet{ladhak2023pre} observed that name-nationality  stereotypes propagate from pretraining data to downstream summarization systems and manifest as hallucinated facts.

\section{Experimental Setup}

Considering the extensive literature on fairness in natural language processing, which highlights significant disparities in the processing of data from different social groups, whether along the dimensions of gender or race or others, we are compelled to ask two questions: 
\begin{enumerate}
   
\item What happens when the input data to be summarized is deliberately grouped according to the social groups, such as dialect groups in our case? (in Section~\ref{sec:position}) and, 

\item  How do the effects of position bias affect the fairness of generated summaries (Section~\ref{sec:fairness}). 
\end{enumerate}

Before exploring these questions, we first describe our experimental setup in this section.

\subsection{Task Formulation}

Considering a multi-document set of $n$ topically-related documents $\mathcal{D} = \{d_1^{g_1}, ..., d_n^{g_r}\}$, where each document belongs to one of several diverse social groups $\mathcal{G} = \{g_1, ..., g_r\}$, the objective is to produce a summary $\mathcal{S(\mathcal{D})}$ that ideally exhibits both high textual quality and fairness. In this work, because of the original dataset design where the number of documents from each group is equal in the input, our investigation is concerned with the notion of equal representation. As such, a summary is considered to be fair when all groups ${g_1, ..., g_r}$  are equally represented in the output.

\subsection{Dataset}
For our experiments, we use the {DivSumm} dataset\footnote{\url{https://github.com/PortNLP/DivSumm}}, an MDS dataset consisting of English tweets of three diverse dialects ({\em African-American} English, {\em Hispanic-aligned} Language, and {\em White-aligned} Language) \cite{olabisi2022analyzing}, which was developed using a large corpus of tweets originally collected by \citet{blodgett2016demographic}.  The dataset includes 25 topically-related sets of documents (tweets) as input and corresponding human-written extractive and abstractive summaries. Each set $\mathcal{D}$ consists of 90 documents evenly distributed among the three dialects (i.e., 30 documents per dialect). A selection of dialect diverse tweets from DivSumm is presented in Table~\ref{abstractive_summaries}.

\subsection{\texttt{Shuffled} and \texttt{Ordered}}

To study the phenomenon of position bias in social multi-document summarization where documents originate from different social groups, we devise two distinct scenarios:  \texttt{shuffled} and \texttt{ordered}, as depicted in Figure~\ref{fig:Ordered.png}.

In the \texttt{\bf shuffled} setting, documents appear randomly present in the input in no specific order. In fact, to ensure consistency, we retain the original order as presented in the DivSumm dataset which the annotators used to craft the summaries.

In the \texttt{\bf ordered} setting, we perturb the input data by grouping documents from each social group together. When the subset of {\em White-aligned} Language tweets ($\mathcal{D}^w$) appears first, the input set is denoted as \texttt{ordered}\textsuperscript{\texttt{white}} or, simply, $\mathcal{O}^w$. Similarly, when the subset of {\em African-American} English tweets ($\mathcal{D}^a$) come first, we denote that set as $\mathcal{O}^a$, and when the subset of {\em Hispanic-aligned} Language documents  ($\mathcal{D}^h$) appears first, we denote that set as $\mathcal{O}^h$. Specifically, the input documents are ordered as follows:
%$\mathcal{O}^{w} = \{d^w_{1,...,30}, d^a_{31,...,60}, d^h_{61,...,90}\}$
%$\mathcal{O}^{a} = \{d^a_{1,...,30}, d^h_{31,...,60}, d^w_{61,...,90}\}$
%$\mathcal{O}^{h} = \{d^h_{1,...,30}, d^w_{31,...,60}, d^a_{61,...,90}\}$
\begin{align*}
\mathcal{O}^{w} = \{\mathcal{D}^w, \mathcal{D}^a, \mathcal{D}^h\} \\
\mathcal{O}^{a} = \{\mathcal{D}^a, \mathcal{D}^h, \mathcal{D}^w\} \\
\mathcal{O}^{h} = \{\mathcal{D}^h, \mathcal{D}^w, \mathcal{D}^a\}
\end{align*}

%\medskip
\noindent These documents are summarized using several models described in the next section, allowing us to subsequently investigate the different summaries we generate -- $\mathcal{S}(\mathcal{O}^w)$, $\mathcal{S}(\mathcal{O}^a)$, $\mathcal{S}(\mathcal{O}^h)$, and $\mathcal{S}$(\texttt{shuffled}) -- which are obtained from four distinct sets of input documents -- $\mathcal{O}^w$, $\mathcal{O}^a$, $\mathcal{O}^h$, and \texttt{shuffled}, respectively. %The system-generated summaries from each of these are evaluated in terms of both quality and fairness.

\subsection{Summarization Models}
We study a total of seven abstractive models in our experiments. We also study three extractive models, the details and results of which are discussed in \ref{appendix:a}. Following the setup of DivSumm, we generate summaries of 5 sentences per topic

The seven abstractive models included in our experiments are as follows: \begin{itemize}
\item {\textsc{Bart}}\footnote{Model checkpoints for BART, T5, LED, Pegasus, and Primera were accessed from \url{https://huggingface.co/models}.} \cite{lewis2019bart}, \item {\textsc{T5}} \cite{raffel2019exploring}, 
\item {\textsc{LED}} (Longformer Encoder-Decoder) \cite{beltagy2020longformer}, 
\item {\textsc{Pegasus}} \cite{zhang2020pegasus}, 
\item {\textsc{GPT-3.5}}, 
\item {\textsc{Primera}} \cite{xiao2021primera}, and 
\item {\textsc{Claude}} (\texttt{Claude 3 Opus}).
\end{itemize}

%(\texttt{text-davinci-003} model)

GPT-3.5 and Claude were prompted {with the following prompt -- ``Please summarize the following text{s} in only five sentences''.}

\section{Position Bias in Social MDS} \label{sec:position}

 \begin{figure}[t]
 \centering
% \begin{subfigure}[b]{0.4\textwidth}
%         \centering
 \includegraphics[width=0.5\textwidth]{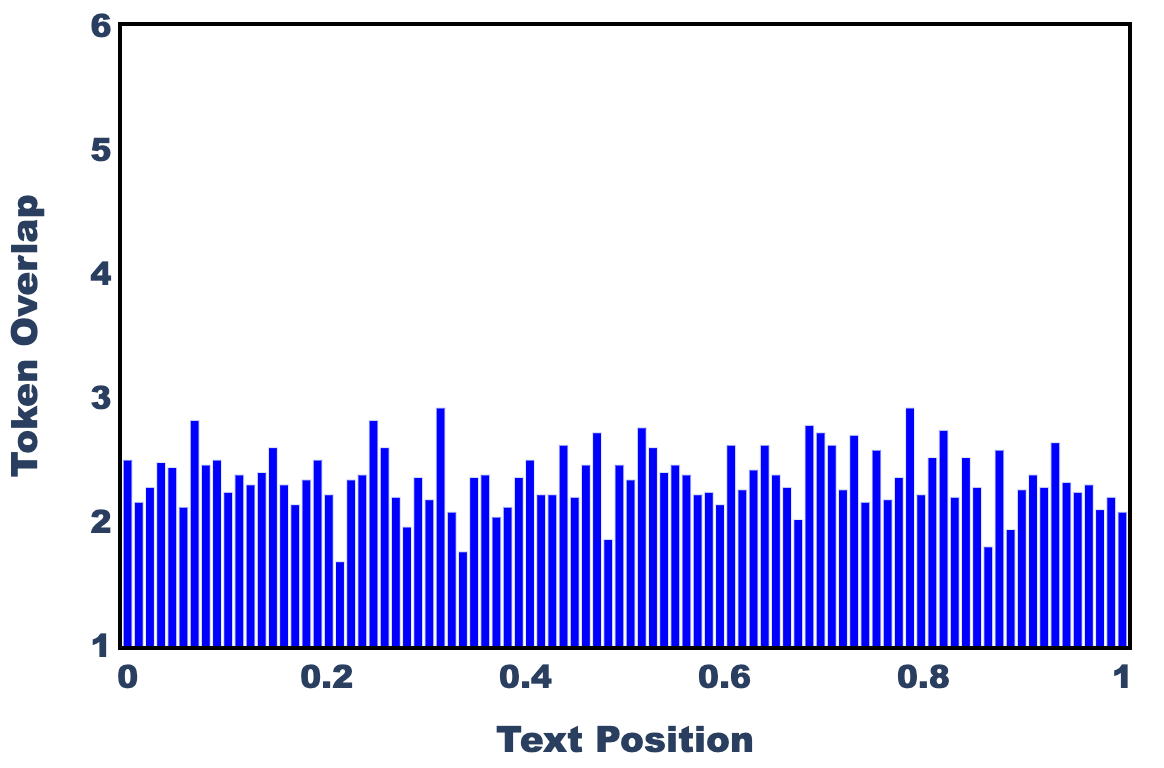}
%          \caption{\emph{Reference summaries (abstractive)}}
%          \label{fig:AbsOverlap}
%     \end{subfigure}
%   \begin{subfigure}[b]{0.4\textwidth}
%          \centering         
%   \includegraphics[width=\textwidth]{img/OverlapDivSummExt.png}
%          \caption{\emph{Reference summaries (extractive)}}
%          \label{fig:ExtOverlap}
%     \end{subfigure}
  
 \caption {{Average token overlap between human-written reference summaries and each document $d_i$ using the DivSumm dataset. Text position on the $x$-axis has been normalized between 0 and 1.}}
  \label{fig:Overlap}
%\vspace{-0.2cm}
\end{figure}

This section discusses position bias within three types of summaries: human-authored reference summaries of the DivSumm dataset, system summaries generated using the \texttt{shuffled} input, and system summaries generated using \texttt{ordered} inputs. Following prior work on position bias, we calculate the overlap  between the  summaries and the input documents {by computing the number of tokens shared between the  summary and each document of the MDS topic set.} That is, given the 90 documents in each topically-related input set, we get the overlap score for each document ($d_1, d_2, ... , d_{90}$) with respect to a summary, and report the average score over the entire dataset. {A higher overlap score implies more semantic relationship between the summary and source document.}

\begin{figure}[t]
%\centering
%\begin{subfigure}[b]{0.4\textwidth}
         \centering
         \includegraphics[width=0.5\textwidth]{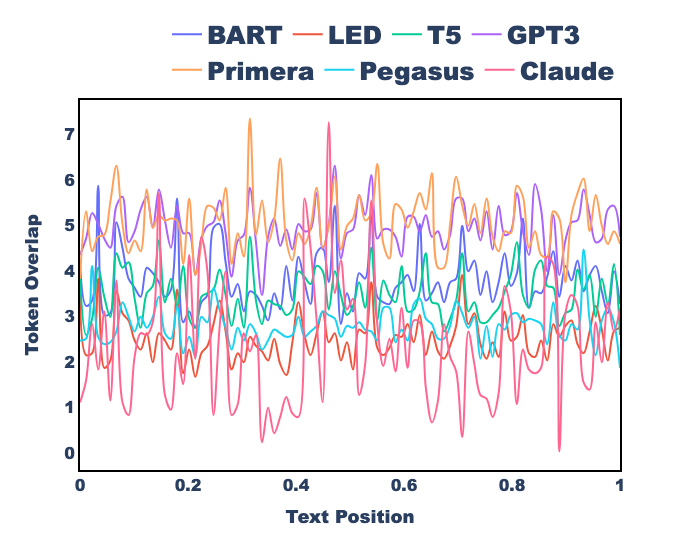}
    %      \caption{\emph{Model summaries (abstractive)}}
    %      \label{fig:AbsModelOverlap}
    % \end{subfigure}
    % \begin{subfigure}[b]{0.4\textwidth}
    %      \centering
    %         \includegraphics[width=\textwidth]{img/ModelOverlapExt.png}
    %      \caption{\emph{Model summaries (extractive)}}
    %      \label{fig:M-ModelOverlap}
    % \end{subfigure}
 
 \caption {{Average token overlap between \texttt{ordered} system-generated summaries by each  abstractive summarization model and each document $d_i$ in the input set $\mathcal{D}$ of DivSumm. Text position on the x-axis has been normalized between 0 and 1.}}
  \label{fig:ModelOverlap}
%\vspace{-0.5cm}
\end{figure}
\subsection{Position Bias in Human-Written Reference Summaries}

To examine position bias in the summaries created by humans, we analyze both abstractive and extractive reference summaries of {DivSumm} dataset.  %This process is repeated for all the topics.
Because the dataset contains two reference summaries per input, we report the average score.  %with respect to both reference summaries and then average the score.
The results are presented in Figure~\ref{fig:Overlap} where no noticeable position bias is observed, and it is encouraging to note that the annotators were not influenced by the position of the documents in the input when producing their summaries. %As expected, higher overlap scores are observed in the case of extractive summaries with an average of 3.98 compared to abstractive summaries with an average of 2.36 tokens. However, 

\subsection{Position Bias in System Summaries (\texttt{Shuffled})}

\begin{figure*}[t]
\centering
  \begin{subfigure}[b]{0.3\textwidth}
         \centering
            \includegraphics[width=\textwidth]{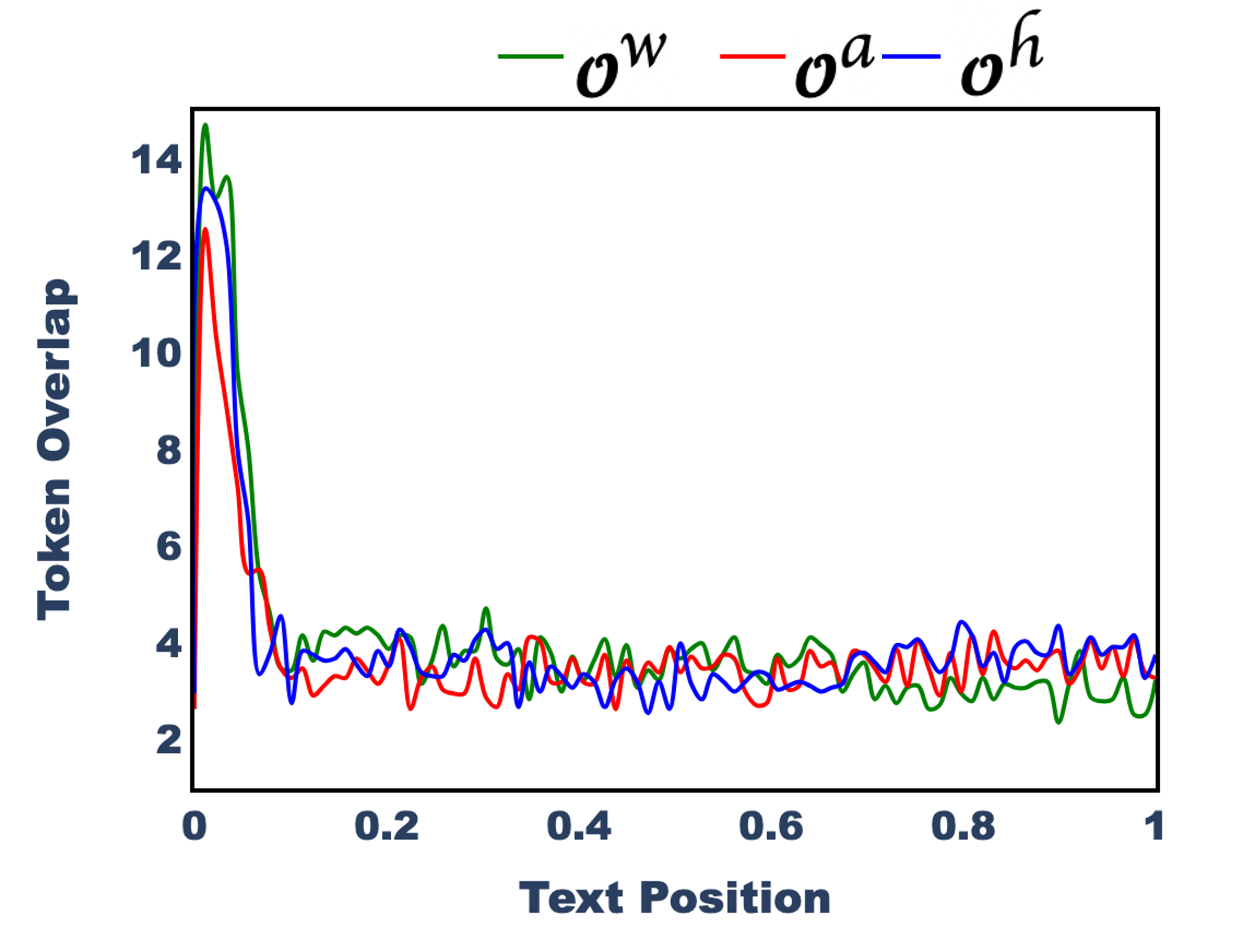}
         \caption{{BART}}
         \label{fig:BARTOverlap}
    \end{subfigure}
    \begin{subfigure}[b]{0.3\textwidth}
         \centering
         \includegraphics[width=\textwidth]{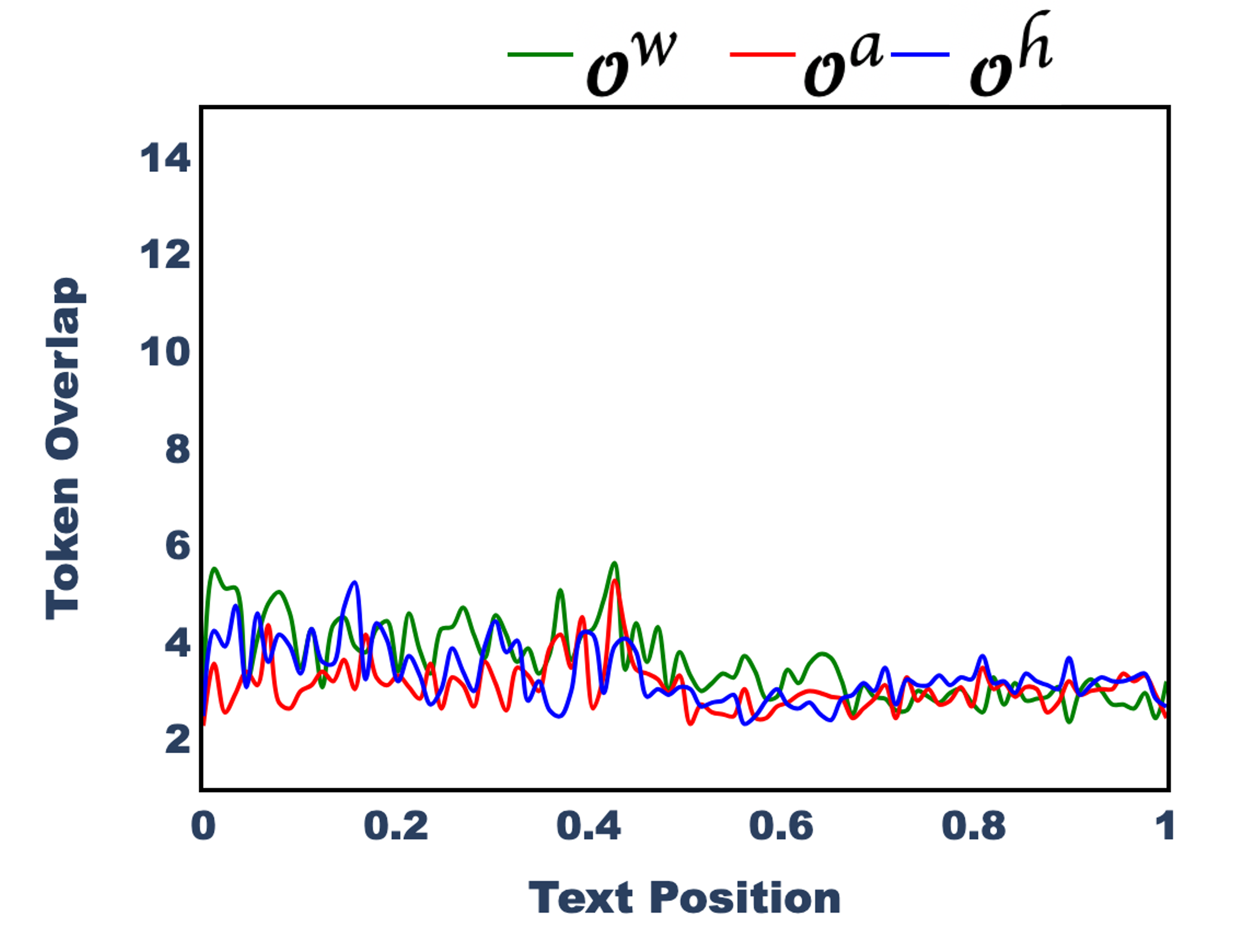}
         \caption{{T5}}
         \label{fig:T5Overlap}
    \end{subfigure}
  \begin{subfigure}[b]{0.3\textwidth}
         \centering
         \includegraphics[width=\textwidth]{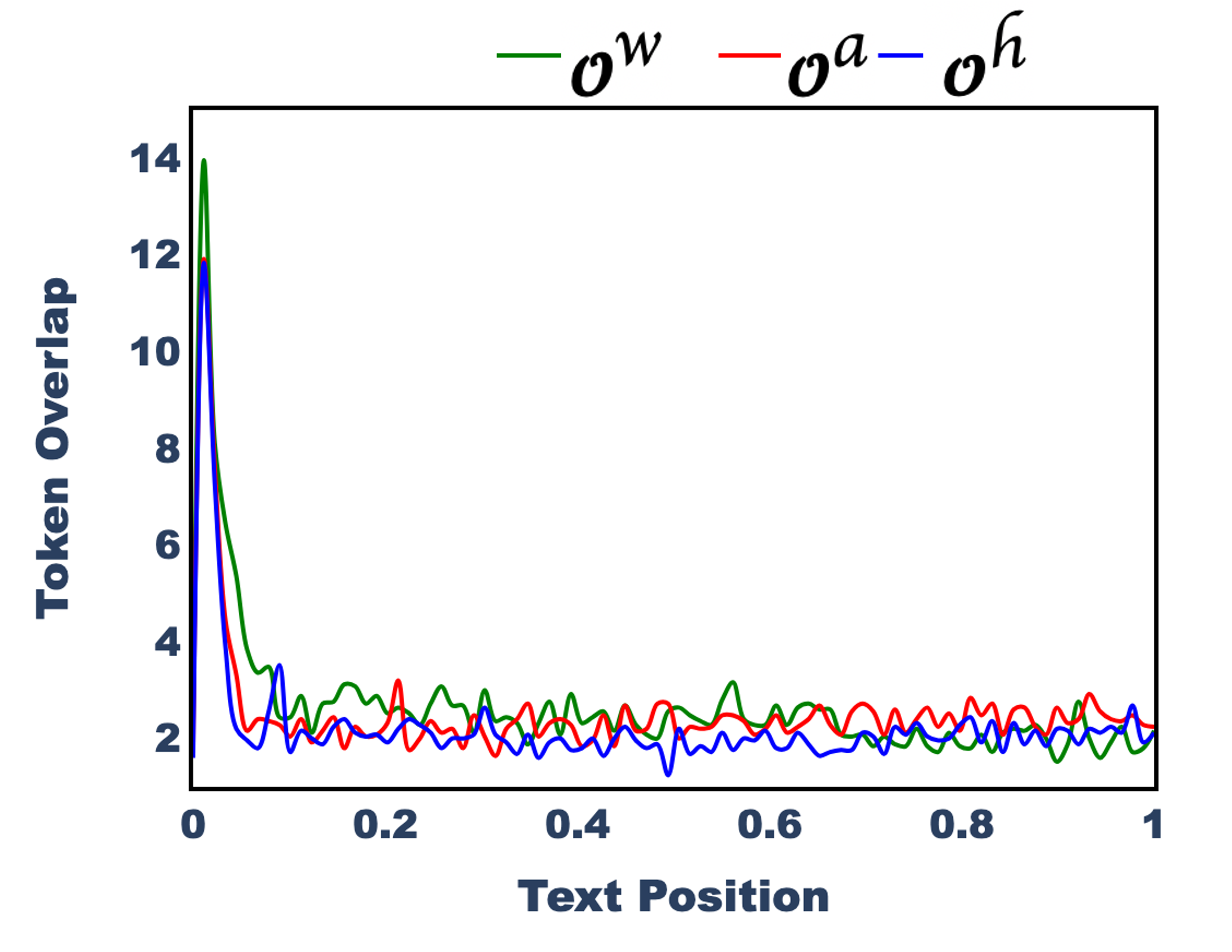}
         \caption{{LED}}
         \label{fig:LEDOverlap}
    \end{subfigure}
  \begin{subfigure}[b]{0.3\textwidth}
         \centering
         \includegraphics[width=\textwidth]{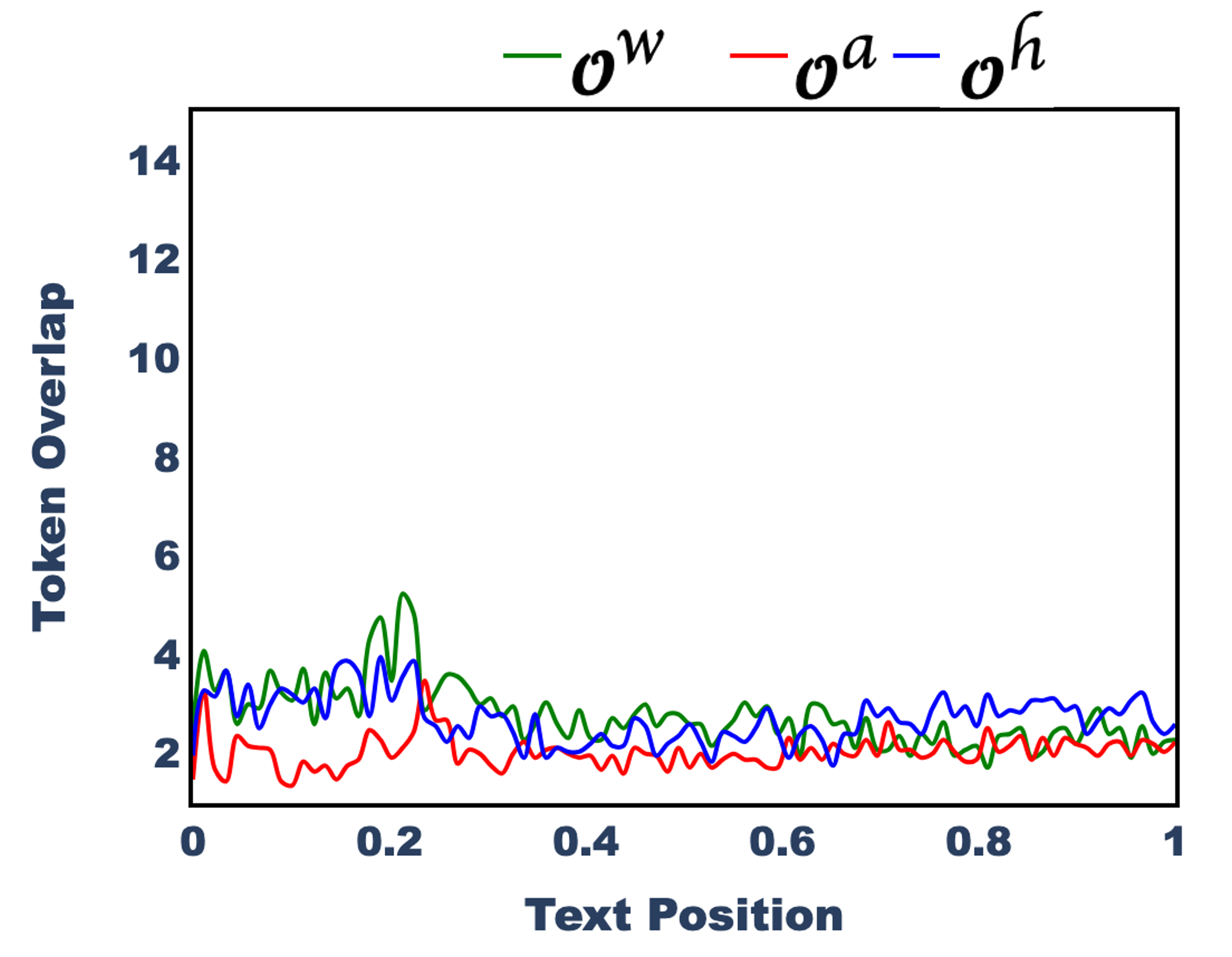}
         \caption{{Pegasus}}
    \label{fig:PegasusOverlap}
    \end{subfigure}
\begin{subfigure}[b]{0.3\textwidth}
         \centering
            \includegraphics[width=\textwidth]{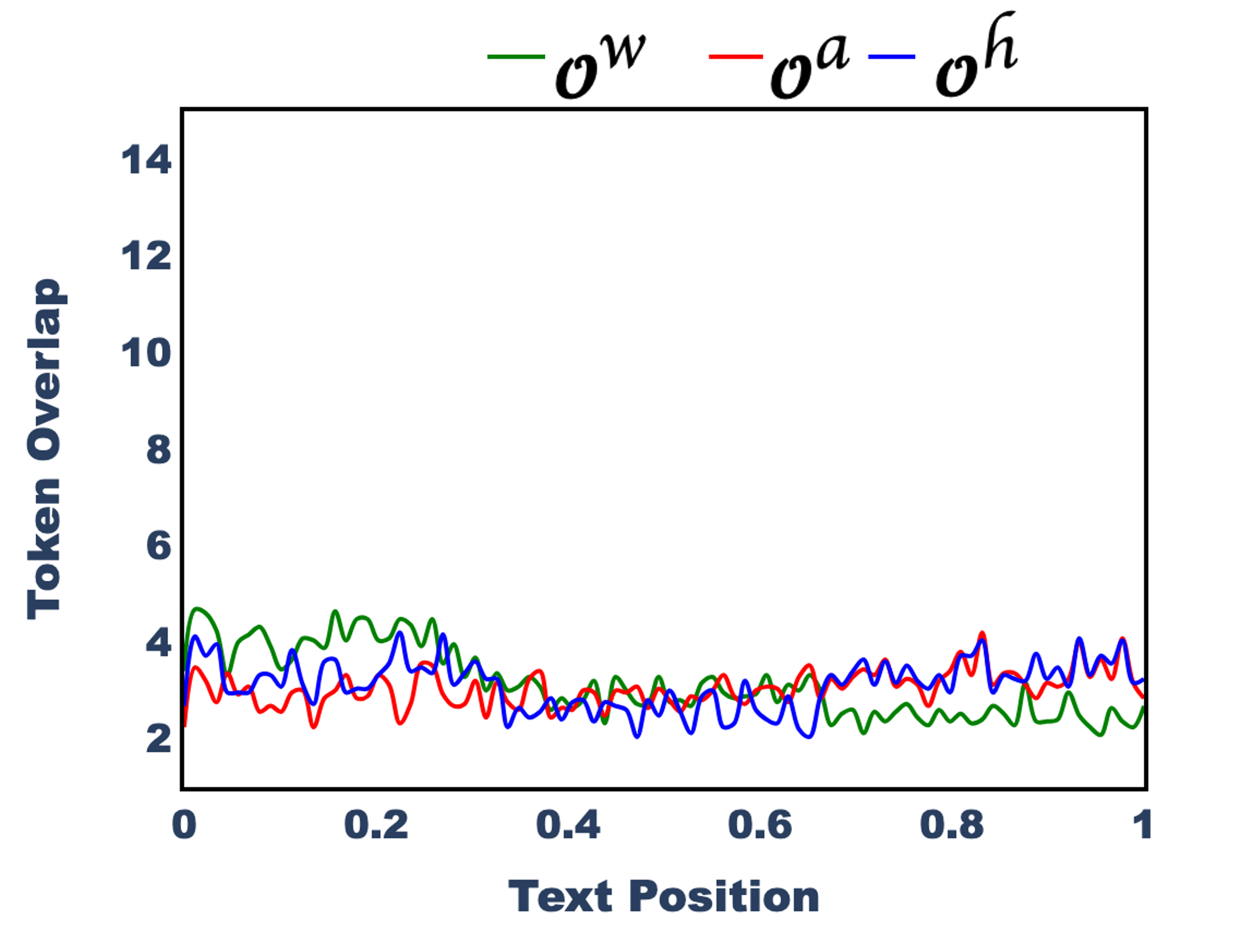}
         \caption{{GPT-3.5}}
         \label{fig:GPT3Overlap}
    \end{subfigure}
  \begin{subfigure}[b]{0.3\textwidth}
         \centering
         \includegraphics[width=\textwidth]{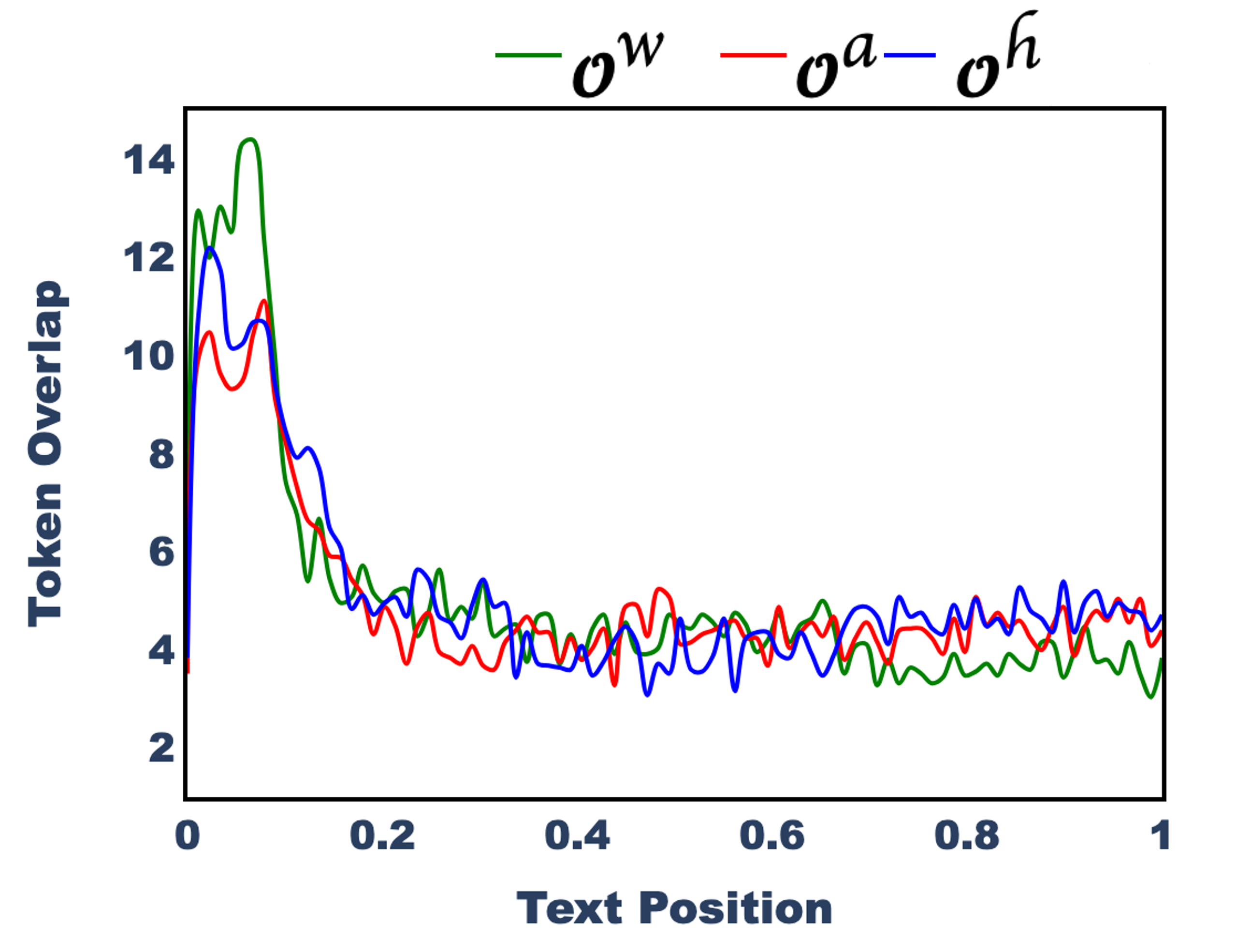}
         \caption{{Primera}}
         \label{fig:PrimeraOverlap}
    \end{subfigure}
\begin{subfigure}[b]{0.3\textwidth}
         \centering
         \includegraphics[width=\textwidth]{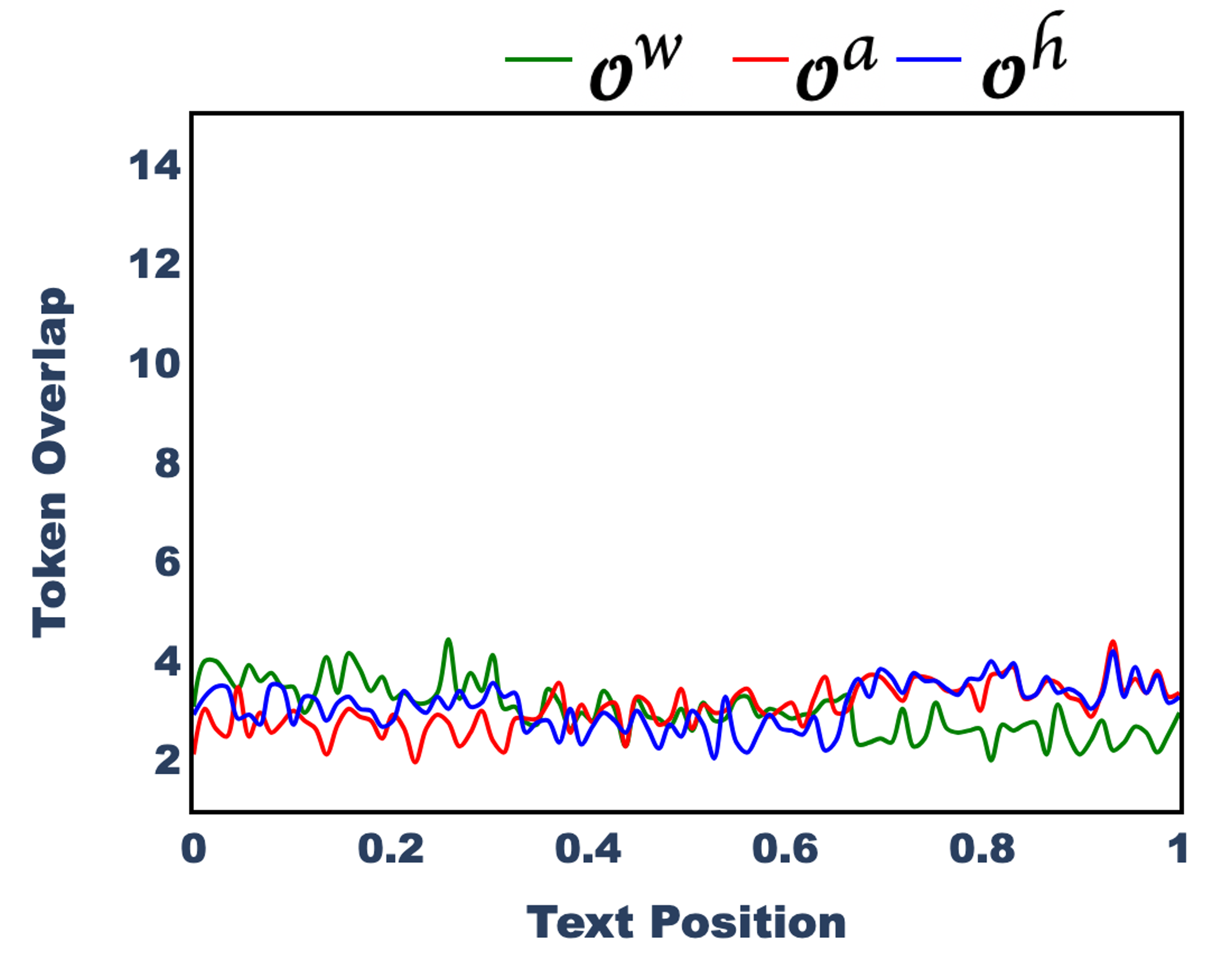}
         \caption{{Claude}}
         \label{fig:ClaudeOverlap}
    \end{subfigure}
 
 \caption {{Average token overlap between \texttt{ordered} system-generated summaries by each of the seven abstractive summarization models and each document $d_i$ in the input set $\mathcal{D}$ of the DivSumm dataset. Text position on the $x$-axis has been normalized between 0 and 1.}}
  \label{fig:OrderedOverlap}
\end{figure*}

The results of position bias within model-generated summaries using \texttt{shuffled} inputs are presented in Figure~\ref{fig:ModelOverlap}. Similar to the human-written reference summaries, we observe no notable position bias suggesting that when summarizing randomly shuffled data from various social groups, the models also do not exhibit any particular lead bias. This observation on DivSumm, a dataset of tweets, is consistent with trends observed in other social datasets (Reddit posts \cite{kim-etal-2019-abstractive} and social user posts \cite{sotudeh-etal-2022-mentsum}).

\subsection{Position Bias in System Summaries (\texttt{Ordered})}
Now we discuss the results of position bias in system summaries that were generated using various \texttt{ordered} inputs: $\mathcal{O}^{w}$, $\mathcal{O}^{a}$, $\mathcal{O}^{h}$. 
%i.e., we count the number of occurrences of each word in summary $\mathcal{S}$({\textbf{$\mathcal{O}^{w}$}}) and each $d_i$ of $\mathcal{O}^w$ for $i=1, 2, .., 90$. We do this for all the 25 topics in the dataset, and plot the token overlap for each input set. 
Model-specific results are presented in Figure~\ref{fig:OrderedOverlap}, where, interestingly, we now observe a {\bf strong position bias in three out of seven abstractive models,} (BART, LED, and Primera), with up to 3 times higher token overlap in the beginning of the input document, as shown by the distribution. Three other models show weak position bias (T5, Pegasus, and GPT-3.5). This phenomenon diverges from traditional position bias, where models tend to favor earlier bits of text. {\em Instead, we notice that models favor earlier pieces of text only when the text exhibits some socially linguistic similarity}. These observations highlight the importance of more nuanced analysis when exploring position bias in summarization systems, especially when processing diverse social data.

\begingroup\tabcolsep=2pt\def\arraystretch{1}
\begin{table*}[t]
\centering
\small
\begin{tabular}{lccca|ccca|ccca|ccca}
\midrule

\multirow{2}{*}{\textbf{Model}} &  \multicolumn{4}{c}{\textbf{$\mathcal{O}^{w}$}} & \multicolumn{4}{c}{\textbf{$\mathcal{O}^{a}$}}  & \multicolumn{4}{c}{\textbf{$\mathcal{O}^{h}$}}  & \multicolumn{4}{c}{\texttt{shuffled}} \\
\cmidrule{2-17}
& {$\mathcal{D}^w$} & {$\mathcal{D}^a$}  & {$\mathcal{D}^h$}  & {$\Delta$Fair ($\downarrow$)} 
& {$\mathcal{D}^w$} & {$\mathcal{D}^a$}  & {$\mathcal{D}^h$} & {$\Delta$Fair ($\downarrow$)} 
& {$\mathcal{D}^w$} & {$\mathcal{D}^a$}  & {$\mathcal{D}^h$} & {$\Delta$Fair ($\downarrow$)} 
& {$\mathcal{D}^w$} & {$\mathcal{D}^a$}  & {$\mathcal{D}^h$} & {$\Delta$Fair ($\downarrow$)}\\
 \midrule
% \multicolumn{17}{c}{\textbf{Similarity}}\\
%\midrule
\textsc{BART} &\textbf{0.64}  &0.41  &0.45  & 0.23 &0.41  &\textbf{0.55}   &0.40  & 0.15 &0.44  &0.42  &\textbf{0.59}  &0.17 &\textbf{0.41} & \textbf{0.41} &0.40  &0.01\\

\textsc{LED} &\textbf{0.47}  &0.30  &0.33  &0.17 &0.31 &\textbf{0.43}   &0.31  &0.12 &0.26  &0.24  &\textbf{0.36} &0.12 &0.30 & 0.29  &\textbf{0.35} &0.06\\

\textsc{T5} &\textbf{0.52}   &0.39  &0.48  &0.13 &0.39  &\textbf{0.46}   &0.43  &0.07 &0.40  &0.41  &\textbf{0.49}  &0.09 & 0.37 & \textbf{0.41} & 0.40 &0.04\\

\textsc{Pegasus} & \textbf{0.34} &0.28 &0.29  &0.06 & 0.22 & \textbf{0.25}  &{0.21}  &0.04 & 0.26 &0.24  & \textbf{0.32} &0.08 & \textbf{0.32} & 0.33 & \textbf{0.32} &0.01\\

\textsc{GPT-3.5} & \textbf{0.47} &0.35 &0.38  &0.12 & \textbf{0.38} & \textbf{0.38}  &0.36  &0.02 & 0.38 &0.34  & \textbf{0.41} &0.07 & \textbf{0.40} & 0.35 & 0.37 &0.05 \\

\textsc{Primera} & \textbf{0.62} &0.41 &0.45  &0.21 & 0.42 & \textbf{0.60}  &0.44  &0.18& 0.45 &0.44  & \textbf{0.62} &0.18& \textbf{0.49} & 0.48 & 0.50 &0.02\\

\textsc{Claude} & \textbf{0.39} &0.33 &0.36  &0.06 & \textbf{0.37} & 0.32  &0.34 &0.05 & \textbf{0.36} &0.31  & 0.34 &0.05 & \textbf{0.37} & 0.32 & 0.35 &0.05\\

 \midrule
 
\textsc {Avg} & {\bf 0.49} & 0.35 & 0.39 & 0.14 & 0.36 & {\bf 0.43} & 0.36 & 0.09 & 0.36 & 0.35 & {\bf 0.45} & 0.11 & 0.38 & 0.37 & 0.39 & 0.04\\

\bottomrule
\end{tabular}
\caption{\textbf{\em Fairness.} Similarity scores of summaries generated by \texttt{ordered} inputs ($\mathcal{O}^{w},\mathcal{O}^{a},\mathcal{O}^{h}$) and \texttt{shuffled} inputs compared to each group of documents ({$\mathcal{D}^w$}, {$\mathcal{D}^a$}, {$\mathcal{D}^h$}) across seven abstractive summarization models using the \emph{DivSumm} dataset. The highest similarity scores are shown in bold.}
\label{tab:results1}
\end{table*}
\endgroup 

\section{Fairness and Textual Quality Amidst Position Bias} \label{sec:fairness}
Having observed an instance of position bias, especially when input data is grouped according to dialect groups, the next natural question to ask is how does this position bias quantitatively impact the fairness and textual quality of the generated summaries. We briefly describe the evaluation metrics before discussing the main results.

\subsection{Evaluation Metrics}
%We consider semantic similarity to estimate fairness (gap) and four metrics to measure the textual quality of the summaries.

\medskip 
\noindent \textbf{\em Fairness (Gap):} \quad One way of measuring fairness is by estimating the amount of representation from each dialect group in the final summary by comparing the summary $\mathcal{S}$ to the set of documents from each group. {Given that an unbiased summary should capture the perspectives across all groups, we evaluate summary fairness for both extractive and abstractive models using semantic similarity of the summary to each represented group.} As an example, for input {\textbf{$\mathcal{O}^{w}$}}, we compare the final summary $\mathcal{S}$({\textbf{$\mathcal{O}^{w}$}}) to the document set of each dialect group: $\mathcal{D}^w$, $\mathcal{D}^w$, and $\mathcal{D}^h$. In other words, we compute $sim(i,j)$ where $i=\{\mathcal{S}(\mathcal{O}^{w}),\mathcal{S}(\mathcal{O}^{a}),\mathcal{S}(\mathcal{O}^{h})\}$ and $j=\{\mathcal{D}^{w},\mathcal{D}^{a},\mathcal{D}^{h}\}$. Similarity can be estimated by many possible methods of obtaining semantic similarity. We use cosine similarity.

From these similarity scores, we can derive the \textbf{Fairness Gap ($\Delta$Fair)} by calculating the difference between the maximum and the minimum scores attributed to any of the groups \cite{olabisi2022analyzing}. %As an example, if the representation scores for the three groups are 80, 70, 50, then the {fairness gap} would be 30. 
Intuitively, a summary that produces relatively similar representation scores across all groups can be considered as {\em fair} because it likely contains comparable representation from all groups such that no one group is significantly underrepresented.

%$\mathcal{D}^w$, $\mathcal{D}^a$, and $\mathcal{D}^h$, 

\medskip
\noindent \textbf{\em Textual Quality:} \quad Four established metrics are used for assessing the quality of the summaries: ROUGE, BARTScore, BERTScore, and UniEval. \textbf{ROUGE} \cite{lin2004rouge} calculates the lexical overlap between the model-generated summary and the reference summaries. For our experiments, we report the F1 scores of ROUGE-L which is the longest common subsequence between the two summaries. \textbf{BARTScore} \cite{NEURIPS2021_e4d2b6e6} leverages BART’s average log-likelihood of generating the evaluated summary conditional on the source document. Since it uses the average log-likelihood for target tokens, the calculated scores are smaller than 0 (negative). We use the \texttt{facebook/bart-large-cnn} checkpoint. \textbf{BERTScore} \cite{bert-score} relies on BERT embeddings and matches words in system-generated summaries and reference summaries to compute token similarity. We use the \texttt{microsoft/deberta-xlarge-mnli} model and report the F1 scores. \textbf{UniEval} \cite{zhong2022towards} is a unified multi-dimensional evaluator that employs boolean question answering format to evaluate text generation tasks. We make use of \texttt{unieval-sum} which evaluates system-generated summaries in terms of four dimensions: coherence, consistency, relevance and fluency. Except for fluency, the rest are reference-free metrics.  We report the overall score.

\begin{comment}
    \textbf{ROUGE} \cite{lin2004rouge} is a reference-based metric used to calculate the lexical overlap between the model-generated summary and the reference summaries. For our experiments, we report the F1 scores of ROUGE-L which is the longest common subsequence between the two summaries. %We average the scores from the comparison of the machine-generated summary to the two reference summaries. 

\textbf{BARTScore} \cite{NEURIPS2021_e4d2b6e6} leverages BART’s average log-likelihood of generating the evaluated summary conditional on the source document. %It uses the weighted log probability of one text given another text, to put different emphasis on different tokens. 
Since it uses the average log-likelihood for target tokens, the calculated scores are smaller than 0 (negative). We use the \texttt{facebook/bart-large-cnn} checkpoint. 

\textbf{BERTScore} \cite{bert-score} relies on BERT embeddings and matches words in system-generated summaries and reference summaries to compute token similarity. We use the \texttt{microsoft/deberta-xlarge-mnli} model.%, and report the F1 scores.

\textbf{UniEval} \cite{zhong2022towards} is a unified multi-dimensional evaluator that employs boolean question answering format to evaluate text generation tasks. We make use of \texttt{unieval-sum} which evaluates system-generated summaries in terms of four dimensions: coherence, consistency, relevance and fluency. Except for fluency, the rest are reference-free metrics.  We report the overall score.
\end{comment}

\subsection{Results}

\begin{figure*}[t]
\centering
  \begin{subfigure}[b]{0.24\textwidth}
         \centering
            \includegraphics[width=\textwidth]{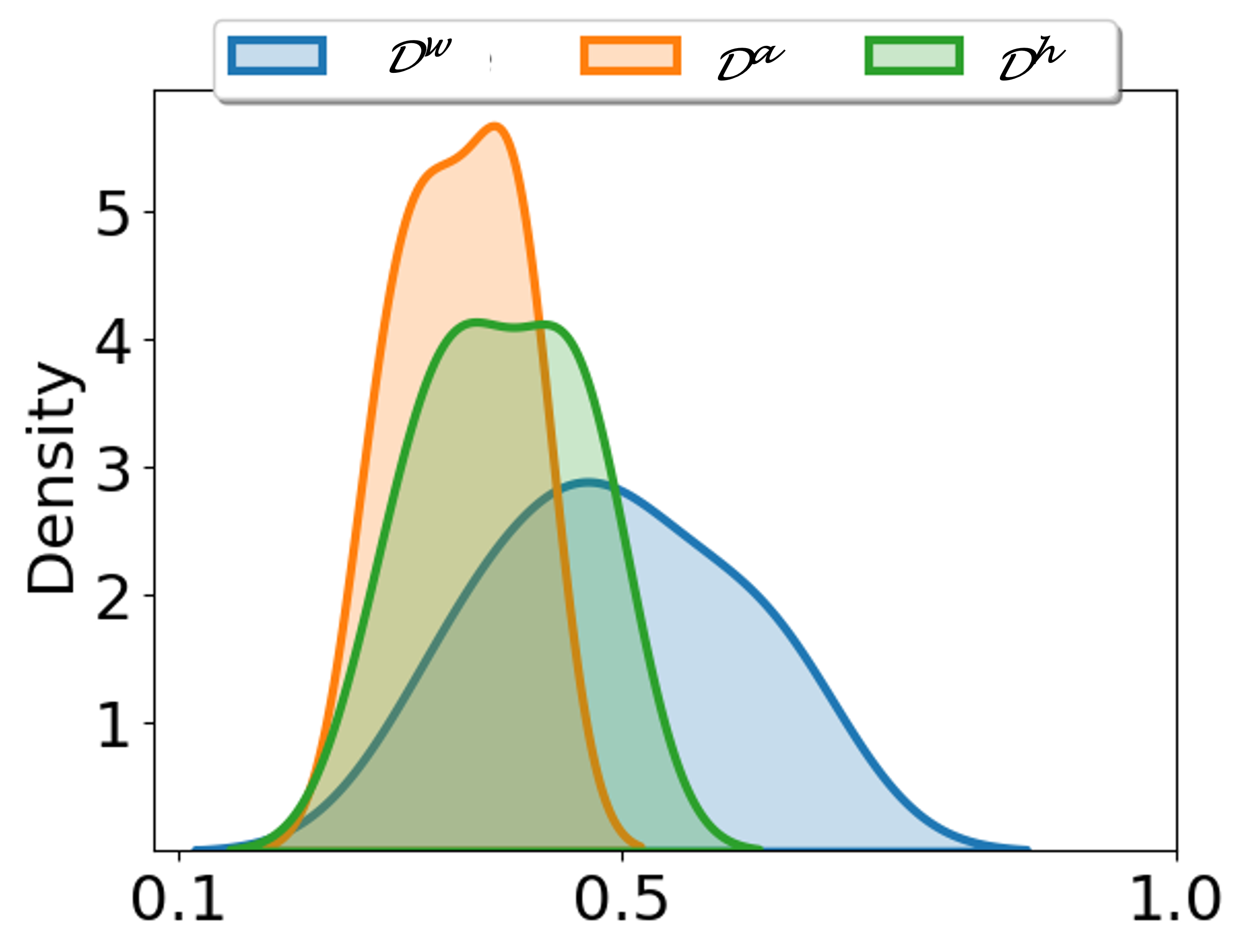}
         \caption{$\mathcal{O}^{w}$}
         \label{fig:OWhite}
    \end{subfigure}
  \begin{subfigure}[b]{0.24\textwidth}
         \centering
         \includegraphics[width=\textwidth]{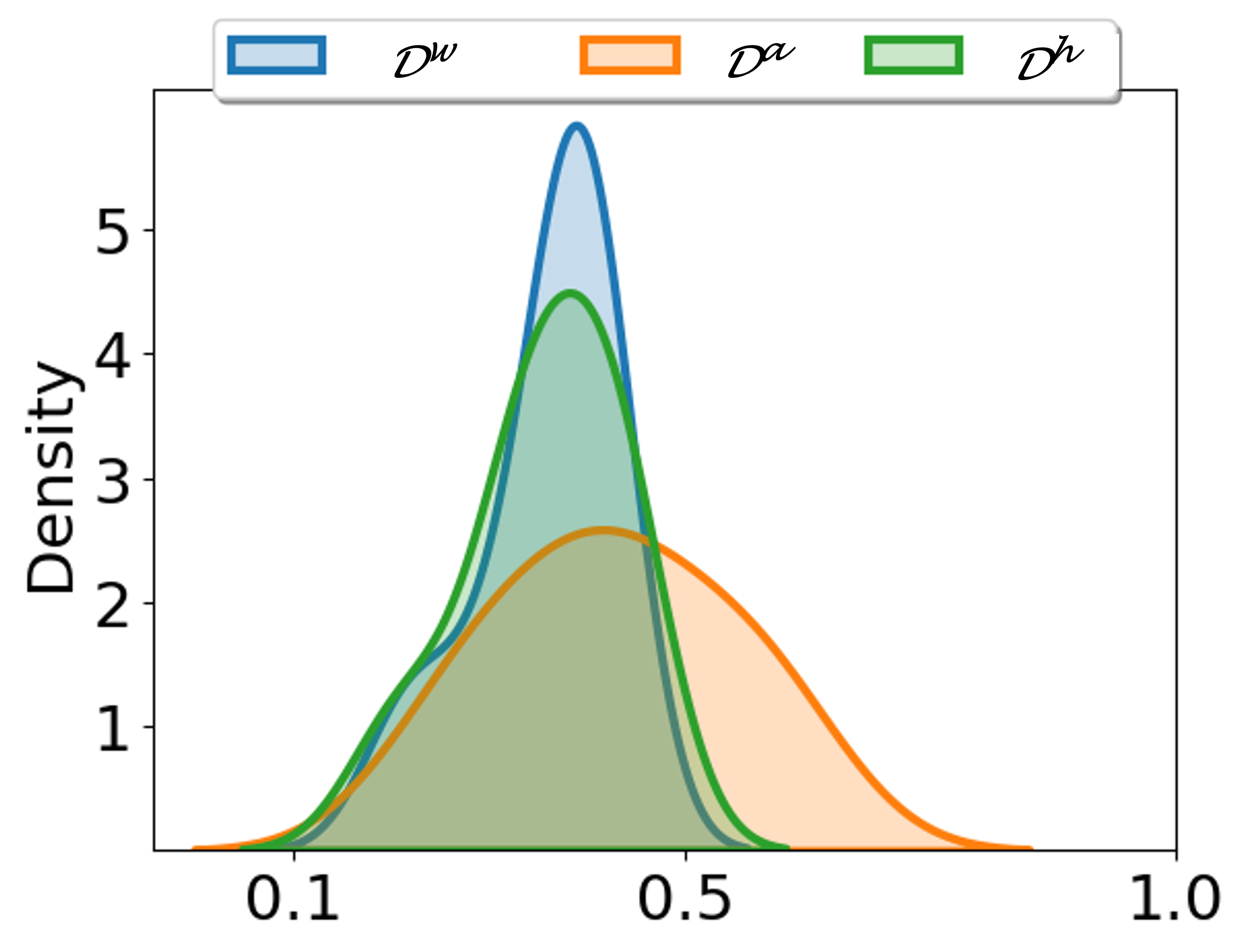}
         \caption{$\mathcal{O}^{a}$}
         \label{fig:OAA}
    \end{subfigure}
  \begin{subfigure}[b]{0.24\textwidth}
         \centering
            \includegraphics[width=\textwidth]{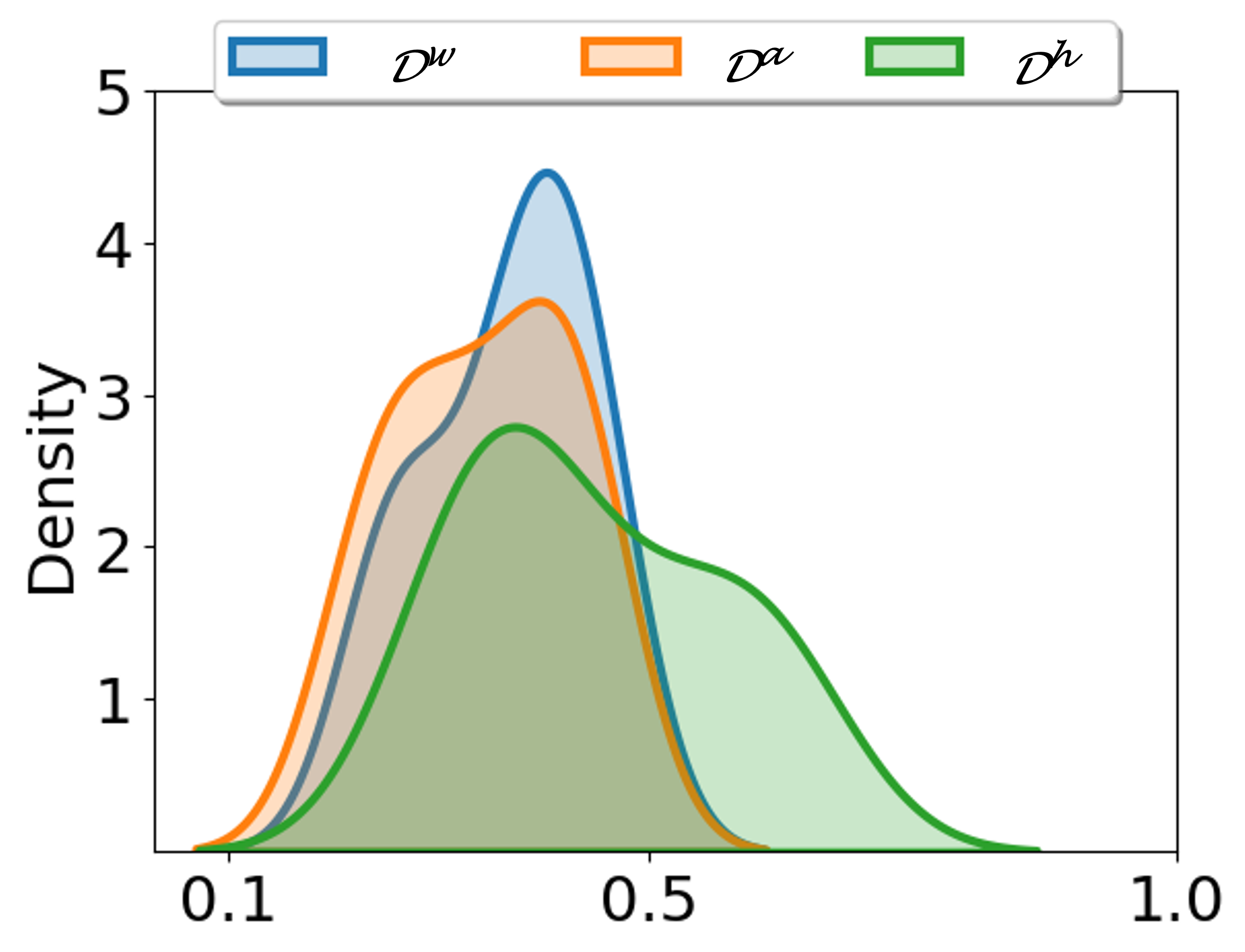}
         \caption{$\mathcal{O}^{h}$}
         \label{fig:OHisp}
    \end{subfigure}
  \begin{subfigure}[b]{0.24\textwidth}
         \centering
            \includegraphics[width=\textwidth]{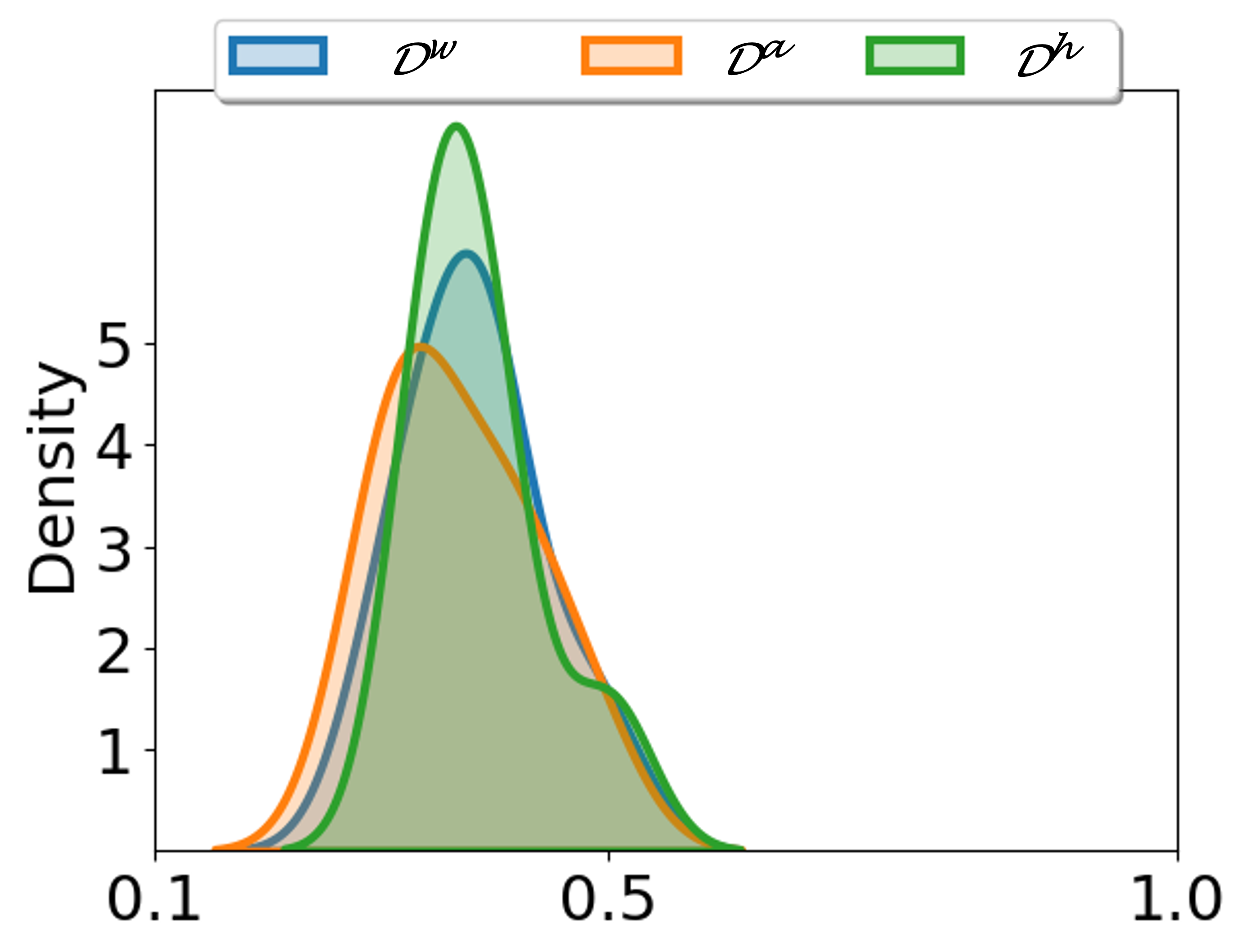}
         \caption{\texttt{shuffled}}
         \label{fig:shuffled}
    \end{subfigure}
 \caption {Density distribution of similarity scores between {system-generated} summaries and each group, across all summarization models for $\mathcal{O}^{w}$, $\mathcal{O}^{a}$, $\mathcal{O}^{h}$ and \texttt{shuffled} input sets. The outputs of \texttt{shuffled} inputs show very different and balanced distributions compared to the \texttt{ordered} inputs.}
  \label{fig:DensityPlots}
%\vspace{0.4cm}
\end{figure*}

\begingroup\tabcolsep=3.5pt\def\arraystretch{1}
\begin{table*}[t]
\centering
\small
\begin{tabular}{lcccc|cccc|cccc|cccc}
\midrule

\multirow{2}{*}{\textbf{Model}} &  \multicolumn{4}{c}{\textbf{{ROUGE-L}} } & \multicolumn{4}{c}{\textbf{\textsc{BARTScore}} } & \multicolumn{4}{c}{\textbf{\textsc{BERTScore}} } & \multicolumn{4}{c}{\textbf{\textsc{UniEval}} }\\
\cmidrule{2-17}
&$\mathcal{O}^{w}$ & $\mathcal{O}^{a}$  & $\mathcal{O}^{h}$  & \texttt{Sh.} 
&$\mathcal{O}^{w}$ & $\mathcal{O}^{a}$  & $\mathcal{O}^{h}$  & \texttt{Sh.}  
&$\mathcal{O}^{w}$ & $\mathcal{O}^{a}$  & $\mathcal{O}^{h}$  & \texttt{Sh.} 
&$\mathcal{O}^{w}$ & $\mathcal{O}^{a}$  & $\mathcal{O}^{h}$  & \texttt{Sh.}\\
 \midrule

\textsc{BART} &\textbf{0.15}  &0.14  &0.14  &\textbf{0.15}  &-3.73  &-3.74  &-3.72  &\textbf{-3.69}  &\textbf{0.51}  &0.50  &\textbf{0.51} &0.50 &0.46 &0.46 &\textbf{0.48} &0.44  \\

\textsc{T5} &\textbf{0.15}  &0.13  &0.13  &0.14  &-3.76  &-3.75  &-3.74  &\textbf{-3.72}  &0.50  &0.48  &0.49  &\textbf{0.51} &0.45 &0.45 &\textbf{0.47} &0.44  \\

\textsc{LED} &\textbf{0.12} &0.11  &0.10  &\textbf{0.12}  &-3.75  &-3.79  &-3.79 &\textbf{-3.73}  &0.44  &0.40  &0.39  &\textbf{0.47} &0.44 &0.44 &\textbf{0.46} &0.43  \\

\textsc{Pegasus} &\textbf{0.14}  &0.11  &0.13  &\textbf{0.14}  &\textbf{-3.73}  &-3.75  &-3.76  &\textbf{-3.73}  &\textbf{0.47}  &0.44  &0.46 &0.46 &0.45 &0.45 &\textbf{0.47} &0.43  \\

\textsc{GPT-3.5} & 0.20 &0.20 & \underline{\textbf{0.21}}  & \underline{\textbf{0.21}} & -3.64 & -3.68  & \underline{\textbf{-3.62}} & -3.65 & 0.57 &0.58  & \underline{\textbf{0.59}} & \underline{\textbf{0.59}} &0.46 &0.45 &\underline{\bf 0.48} &0.44 \\

\textsc{Primera} &\textbf{0.14}  &0.12  &0.13  &0.13  &-3.67  &-3.68  &\textbf{-3.63}  &-3.64  &0.51  &0.49  &\textbf{0.50}  &0.49 &0.45 &0.46 &\underline{\bf 0.48} &0.44  \\

\textsc{Claude} &{0.18}  &0.18  &\textbf{0.19} &0.18  &\textbf{-3.64}  &\textbf{-3.64}  &\textbf{-3.64}  &-3.65  &0.56  &0.56  &\textbf{0.57}  &0.56 &0.44 &0.44 &\textbf {0.46} &0.43  \\

\midrule

\textsc{Avg} & 0.15 & 0.14 & 0.15 & 0.15 & -3.70 & -3.72 & -3.70 & -3.69 & 0.51 & 0.49 & 0.50 & 0.51 & 0.45 & 0.45 & 0.47 & 0.44\\

\bottomrule
\end{tabular}
\caption{\textbf{\em Quality}. Results of \texttt{ordered} {($\mathcal{O}^{w}$, $\mathcal{O}^{a}$, $\mathcal{O}^{h}$)} and \texttt{shuffled} {(\texttt{Sh.})} approaches across seven abstractive summarization models showing ROUGE-L, BARTScore, BERTScore, and UniEval scores on the \emph{DivSumm} dataset. The best scores are shown in {\bf bold}, whereas the highest scores per metric are shown as \underline{underlined}.}
\label{tab:results3}
\end{table*}
\endgroup

\noindent \textbf{\em Evaluating fairness}. The results in Table~\ref{tab:results1} report the fairness scores for all seven models. {\bf We clearly observe that ordering the input documents based on groups certainly favors the group that appears first.} This phenomenon is consistently observed in all three types of \texttt{ordered} sets, regardless of which  particular dialect group's data is presented first. However, when the documents are presented as \texttt{shuffled}, no single group is over-represented and the summaries appear more balanced ($\Delta$Fair = 0.04).

The density plots in Figure~\ref{fig:DensityPlots} also show that the \texttt{shuffled} input set is the most balanced across all groups, unlike the \texttt{ordered} sets which are significantly skewed. Furthermore, amongst \texttt{ordered} documents, the fairness gap is the largest when documents of White-aligned language are passed first ($\Delta$Fair = 0.14), and the smallest when documents of African-American English appear first ($\Delta$Fair = 0.09).

\medskip
\noindent \textbf{\em Evaluating textual quality}. \quad Table~\ref{tab:results3} presents the summary quality scores across all seven summarization models for the four sets of input. We clearly see that the scores of the \texttt{shuffled} approach are superior or comparable to the scores from the three input sets in the \texttt{ordered} approach, except in the case of UniEval. {\bf This shows that with respect to quality, there is no significant difference whether documents are presented as ordered or shuffled.}

\subsection{Discussion}

Some samples of system summaries are presented in Table~\ref{abstractive_summaries}. The key findings of our study can be summarized as follows:

\begin{itemize}
\item We find no evidence of position bias in human-annotated reference summaries of DivSumm, a social MDS dataset of diverse groups. Same observation is made for the abstractive system-generated summaries obtained when the input documents are passed in randomly or shuffled. 
\item However, when the input is \texttt{ordered} based on dialect groups, we observe a significant position bias in the system summaries, with the summaries  having higher overlap with the group that appears first in the input document. 
\item \texttt{Ordered} documents  involving different dialects result in summaries that are significantly skewed in terms of fairness, with the group whose data appears first is clearly favored by the models. In contrast, \texttt{shuffled} documents show the least amount of fairness gap.
\item In terms of quality, we observe that for all models and metrics, the scores for \texttt{ordered} and \texttt{shuffled} remain comparable, suggesting that ordering based on diverse groups has no noticeable effect on the quality of system-generated summaries. 
\end{itemize}

\begingroup\tabcolsep=2pt\def\arraystretch{0.9}
\begin{table*}[!ht]
\centering
\small
\begin{tabular}{p{0.1\textwidth}|p{0.8\textwidth}}
\midrule
\multicolumn{2}{c}{\textbf{Input Documents Set}} \\
\midrule 
\multicolumn{2}{l} 
{$d_1$: \textbf{Hispanic} : The Grammys should have come out on Saturday so I won't stay up late today lol} \\
\multicolumn{2}{l} 
{$d_2$: \textbf{AA} : Wasn't it during the Grammys the last time Chris Brown slid Rhianna?} \\
\multicolumn{2}{l} 
{$d_3$: \textbf{White} : Feel free to join my lonesome self swimminngg at Grammys!!} \\
\multicolumn{2}{l}
{$d_4$: \textbf{AA} : I've given up \#DowntonAbbey for J.T.? This is serious \#Grammys} \\
\multicolumn{2}{l}
{$d_5$: \textbf{Hispanic} : oh lol thanks thought you were talking about the Grammys lol sorry lol} \\
\multicolumn{2}{l}
{$d_6$: \textbf{Hispanic} : I don't even know if I am watching the right latin Grammys lol} \\
\multicolumn{2}{l}
{$d_7$: \textbf{White} : "If I'm a hipster about anything, it's Kings of Leon. I listened to them before they won Grammys."} \\
\multicolumn{2}{l}
{$d_8$: \textbf{White} : isn't performing at the Grammys? What's the point of even having the Grammys now?} \\
\multicolumn{2}{l}
{$d_9$: \textbf{AA} : Imma get some ideas for Easter from these here Grammys . \_\_\_\_} \\
\multicolumn{2}{l}
{$d_{10}$: \textbf{Hispanic} : Wow every celeb in the crowd is singing Bruno's lyrics. Awesome. \#Grammys} \\
\multicolumn{2}{l}
{...} \\
\multicolumn{2}{l}
{...} \\
\multicolumn{2}{l}
{...} \\
\multicolumn{2}{l}
{$d_{88}$: \textbf{AA} : These Grammys need some life in it cause I'm so bored} \\
\multicolumn{2}{l}
{$d_{89}$: \textbf{AA} : Got my easter outfit nd dress for the Grammys.. im set :)} \\
\multicolumn{2}{l}
{$d_{90}$: \textbf{White} : "great, now she can have practice for whe she doesn't win at next year's Grammys\!"} \\

\midrule
\textbf{Model} & \textbf{Summary}\\
\midrule
{\texttt{BART}} & {I've given up DowntonAbbey for J.T.? This is serious Grammys. The Grammys lasted an extra 30 mins just for us to all get hazed by LL smh. Im not even watching the Grammys im too lazy to change the channel but isn't Justin doing a livestream or something at the same time. I'm gonna say yes.}\\

\midrule
{\texttt{GPT-3.5}} & Tina Campbell won't be attending the Grammys, and J Cole is being tipped for success. The Black Keys won five awards at the Grammys, and Eminem is expected to win Best Male Singer and Rapper. Rihanna and Chris Brown were together at the Grammys, and there is a lot of Twitter beef happening due to the event. An ad agency has the Grammys as their client, and a speech teacher asked someone to dress for the Grammys for a show. People are discussing the Grammys and celebrities attending. \\
\bottomrule
\end{tabular}
\caption{Example of a \texttt{shuffled} input set from the \emph{DivSumm} dataset showing input documents with their respective dialects, and system summaries generated by the model with the lowest fairness gap (BART) and highest textual quality (GPT-3.5). Note that the dialect labels shown here are only for clearer presentation and are not part of the input to the model.}
\label{abstractive_summaries}
\end{table*}
\endgroup

Taken together, the findings of our study indicate that both the \texttt{ordered} and \texttt{shuffled} approaches yield comparable results in terms of textual quality, but highly disparate results in terms of fairness. This phenomenon is consistently observed in all abstractive models, suggesting that the models are not robust to fairly straightforward group-level data perturbations. These findings are important because they highlight a potential source of nuanced bias in the summarization models. The observation that ordering the input documents based on groups favors the group that appears first indicates a systematic bias in the models' behavior. The fact that the shuffled input set leads to more balanced summaries across all groups implies that the bias observed in the ordered sets can be mitigated by introducing randomness in the presentation of input data. This insight is crucial for understanding and addressing bias in summarization systems, especially in scenarios where fairness and equity are important considerations, such as in social data analysis or decision-making processes. Overall, this result sheds light on an important aspect of model behavior and informs strategies for improving the fairness and effectiveness of summarization models.

\section{Conclusion}

In this work, we investigate how position bias manifests in social multi-document summarization, specifically in scenarios where the input  data is derived from three linguistically diverse communities. When presented with randomly shuffled input data,  summaries generated by ten distinct summarization models exhibited no signs of position bias. However, a significant shift occurred when the input data was simply reordered based on social groups. In such instances, the models produced biased summaries, primarily favoring the social group that appeared earlier in the input sequence.  In terms of the quality of generated summaries, however, there was no notable difference due to the order in which source documents were presented, whether shuffled or ordered. Our results suggest that position bias manifests differently in the context of social multi-document summarization. Furthermore, they highlight the need to incorporate randomized shuffling in multi-document summarization datasets particularly when summarizing documents from diverse groups to ensure that the resultant summaries are not only of high quality but also faithfully representative of the diversity present in the input data.

\section*{Ethical Considerations}

Our findings and conclusions in this paper are based on an existing social media summarization dataset composed of tweets in English, primarily due to the lack of appropriate resources available to undertake such studies. Given the nature of naturally occurring data, it is possible that the data  contains some offensive language. Hence, it is possible for the models to also generate summaries with offensive words. In addition to this, due to the constraint on tweet length, users are known to use acronyms and slangs that may have various meanings across different groups -- this phenomenon is not accounted for in this study. Also, the existing dataset that we use in this work was originally collected from a corpus using geolocation and census data. This dialectal information used in categorizing users' languages should not be used as a representation of users' racial information. In this work, we evaluate summary fairness using proxy metrics such as semantic similarity to each represented group. The definition of fairness may vary for humans, and as such this should not be used as the gold standard. 

\section*{Acknowledgments}
We thank the anonymous reviewers as well as the members of PortNLP lab for their insightful comments. This research was supported by National Science Foundation grants (CRII:RI 2246174 and SAI-P 2228783).

\bibliography{main}

\appendix

\newpage
\section{Fairness in Extractive Models}
\label{appendix:a}

We repeat the same experiments and analysis for extractive models to observe if they exhibit behavior similar to that observed in the abstractive models.

\subsection{Summarization Systems}
We study three summarization models in our experiments to generate summaries of 5 sentences per topic (multi-document set):

\smallskip

\textbf{\textsc{TextRank}}\footnote{\url{https://radimrehurek.com/gensim_3.8.3/summarization/summariser.html}} \cite{mihalcea2004textrank}, an unsupervised graph-based ranking method, determines the most important sentences in a document based on information extracted from the document itself.% and therefore performs well even without any form of domain knowledge or pretraining.

\textsc{\textbf{Bert-Ext}}\footnote{\url{https://pypi.org/project/bert-extractive-summarizer/}} \cite{miller2019leveraging}, an extractive summarization model built on top of BERT \cite{devlin2018bert}, uses $k$-means clustering to select  sentences closest to the centroid as the summaries. 

\textbf{\textsc{Longformer}}\footnote{Model checkpoint for Longformer was accessed from \url{https://huggingface.co/models}} \cite{beltagy2020longformer} is a modification of the transformer architecture, using a self-attention operation that scales linearly with the sequence length. %This was introduced because despite the success of transformer models in achieving SOTA performance on a wide range of NLP tasks, their ability to efficiently scale on longer inputs remains limited. 

\subsection{Evaluation Metrics}
In evaluating textual quality, We use the same four metrics used for the abstractive models. To estimate fairness (gap), in addition to semantic similarity used in evaluating the fairness of abstractive models, we consider {\bf coverage} as well which measures the extent to which a summary is a derivative of the input text. Following previous literature \cite{dash2019summarizing,keswani2021dialect}, we estimate group fairness via disparity in {\em extractive fragment coverage} \cite{grusky2018newsroom}, which indicates the degree of surface-level text overlap by computing the percentage of words in the summary from each dialect group's collection of documents.

\subsection{Results}
\begin{figure}[!t]
    \centering
    \includegraphics[width=0.45\textwidth]{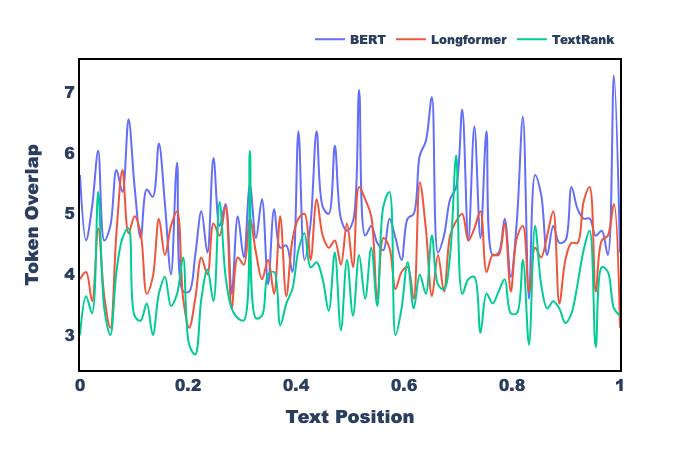}
    \caption {{Average token overlap between \texttt{shuffled} system-generated summaries by each of the three extractive summarization models and each document $d_i$ in the input set $\mathcal{D}$ of DivSumm. Text position on the x-axis has been normalized between 0 and 1.}}
  \label{fig:M-Overlap}
%\vspace{-0.3cm}
\end{figure}

\begin{figure*}[t]
\centering
\begin{subfigure}[b]{0.3\textwidth}
         \centering
         \includegraphics[width=\textwidth]{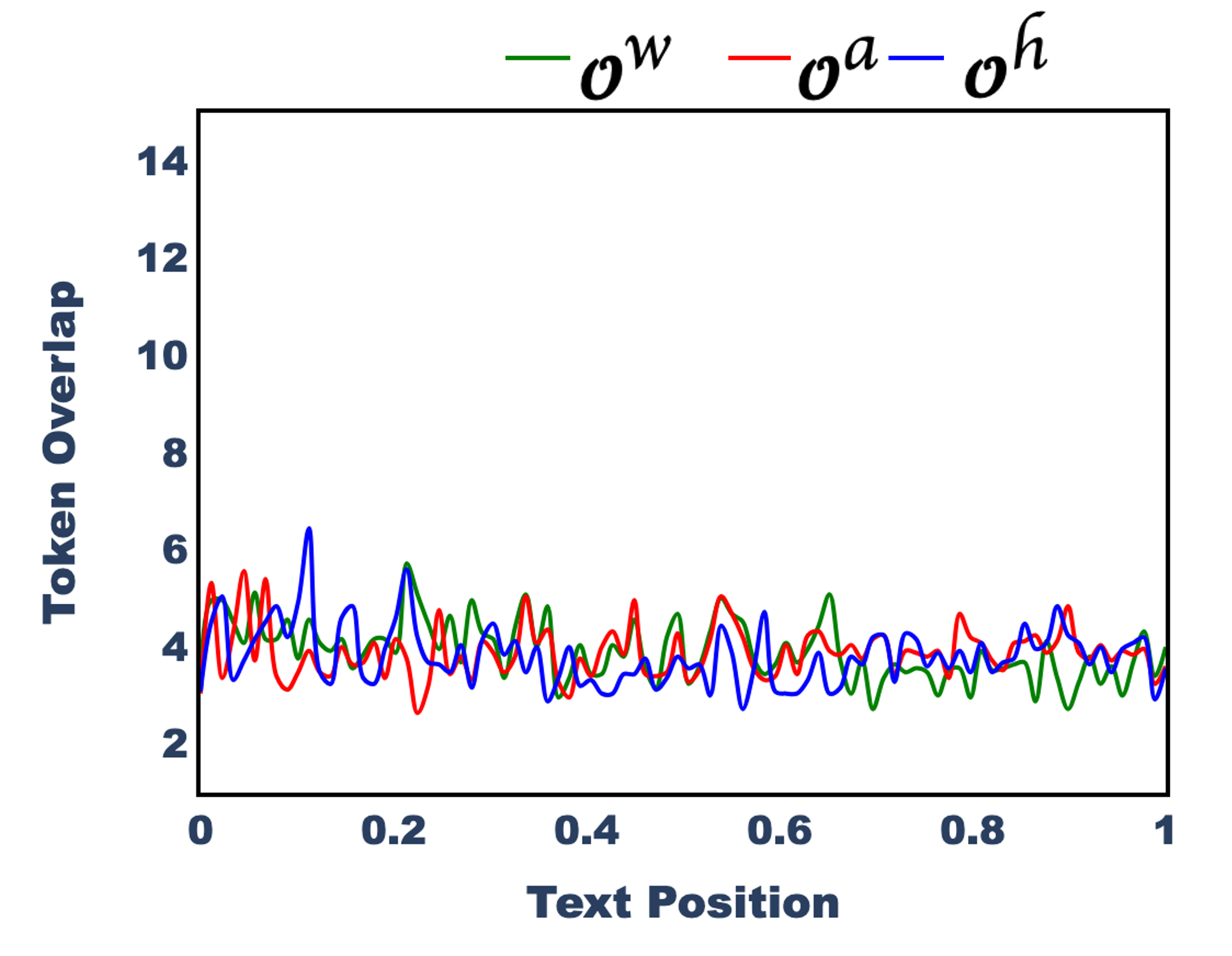}
         \caption{\emph{TextRank}}
         \label{fig:TextrankOverlap}
    \end{subfigure}
    \begin{subfigure}[b]{0.3\textwidth}
         \centering
            \includegraphics[width=\textwidth]{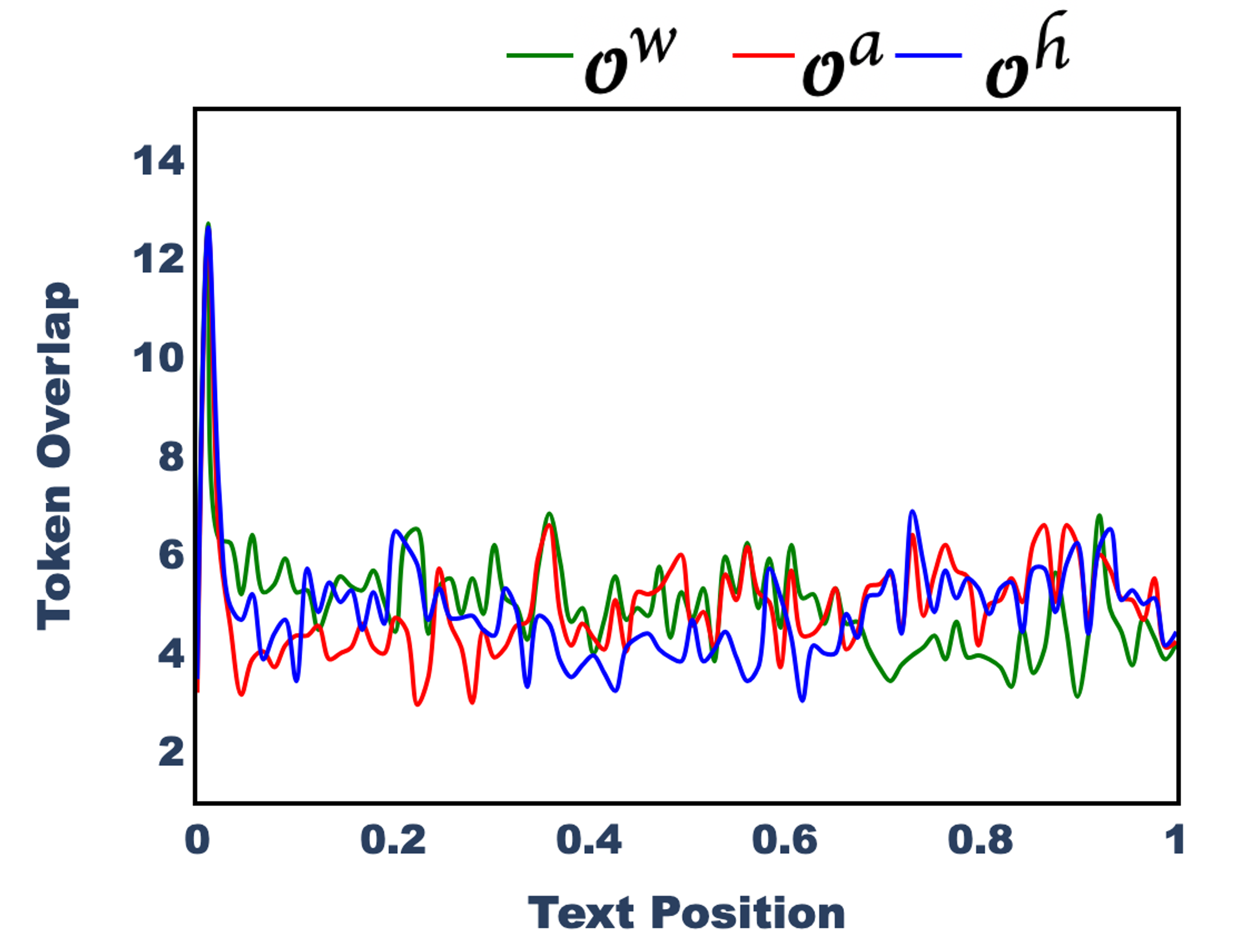}
         \caption{\emph{BERT}}
         \label{fig:BERTOverlap}
    \end{subfigure}
  \begin{subfigure}[b]{0.3\textwidth}
         \centering
         \includegraphics[width=\textwidth]{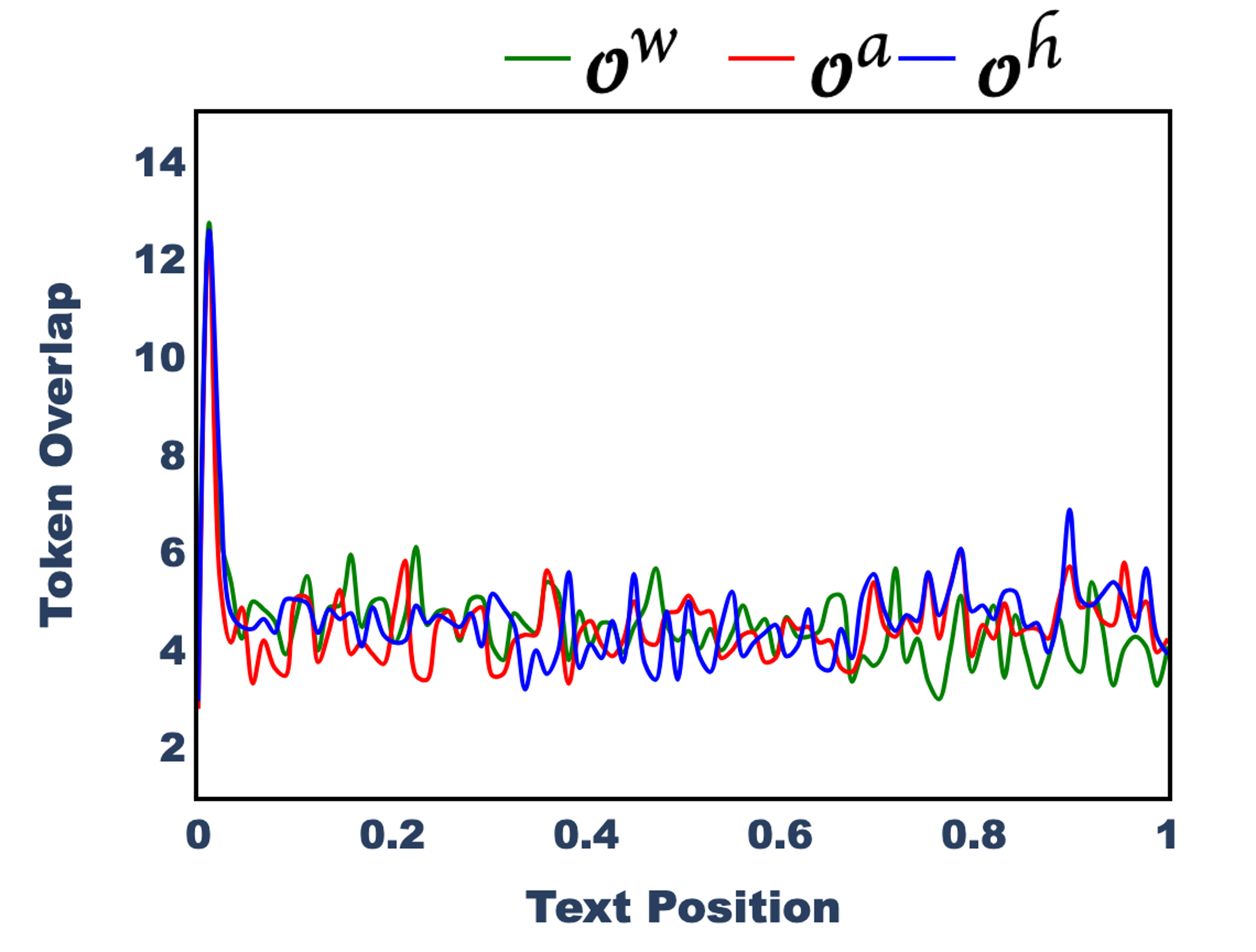}
         \caption{\emph{LongFormer}}
         \label{fig:LongformerOverlap}
    \end{subfigure}

 \caption {{Average token overlap between \texttt{ordered} system-generated summaries by each of the extractive summarization models and each document $d_i$ in the input set $\mathcal{D}$ of DivSumm. Text position on the x-axis has been normalized between 0 and 1.}}
  \label{fig:ExtOrderedOverlap}
%\vspace{-0.5cm}
\end{figure*}

While \texttt{shuffled} extractive models show no noticeable position bias in Figure~\ref{fig:M-Overlap}, we observe a strong position bias using \texttt{ordered} inputs in two out of three extractive models (BERT and LongFormer), as shown in Figure~\ref{fig:ExtOrderedOverlap} further highlighting the importance of exploring position bias in summarization of diverse social data. 

%\noindent \textbf{\em Evaluating fairness}.  \quad 
Tables~\ref{tab:extCoverage} and ~\ref{tab:extSimilarity} show the fairness scores in terms of coverage and similarity, respectively, of extractive summaries. For all three models, we observe that the summaries generated using the \texttt{ordered} sets distinctly favor the group that appeared first in the input set of documents, while this phenomenon is absent from the \texttt{shuffled} set, where the results are much more evenly distributed across the three groups for all three models. Table~\ref{tab:extQuality} presents the quality scores along four metrics where, similar to abstractive models, little difference is noted between \texttt{ordered} and \texttt{shuffled} approaches.

%In terms of the models, in general BERT obtains the highest results amongst extractive models.

% \begin{figure}[t]
% \centering
%   \begin{subfigure}[b]{0.4\textwidth}
%          \centering
%             \includegraphics[trim={0 0.8cm 0 1.2cm},clip,width=\textwidth]{img/ModelOverlapExt.png}
%          \caption{\emph{Model summaries (extractive)}}
%          \label{fig:ExtModelOverlap}
%     \end{subfigure}
%  \caption {{Average token overlap between \texttt{shuffled} system-generated summaries by each of the three extractive summarization models and each document $d_i$ in the input set $\mathcal{D}$ of DivSumm. Text position on the x-axis has been normalized between 0 and 1.}}
%   \label{fig:M-Overlap}
% %\vspace{-0.3cm}
% \end{figure}

\begingroup\tabcolsep=3.5pt\def\arraystretch{1}
\begin{table*}[t]
\centering
\small
\begin{tabular}{lccca|ccca|ccca|ccca}
\midrule

\multirow{2}{*}{\textbf{Model}} &  \multicolumn{4}{c}{\textbf{$\mathcal{O}^{w}$}} & \multicolumn{4}{c}{\textbf{$\mathcal{O}^{a}$}}  & \multicolumn{4}{c}{\textbf{$\mathcal{O}^{h}$}}  & \multicolumn{4}{c}{\texttt{shuffled}} \\
\cmidrule{2-17}
& {$\mathcal{D}^w$} & {$\mathcal{D}^a$}  & {$\mathcal{D}^h$}  & {$\Delta$Fair} 
& {$\mathcal{D}^w$} & {$\mathcal{D}^a$}  & {$\mathcal{D}^h$} & {$\Delta$Fair} 
& {$\mathcal{D}^w$} & {$\mathcal{D}^a$}  & {$\mathcal{D}^h$} & {$\Delta$Fair} 
& {$\mathcal{D}^w$} & {$\mathcal{D}^a$}  & {$\mathcal{D}^h$} & {$\Delta$Fair}\\
% \midrule
%\multicolumn{17}{c}{\textbf{Coverage}}\\
\midrule

\textsc{Textrank} &\textbf{0.80} &0.72  &0.76  &0.08 &0.70  &\textbf{0.81} &0.74  &0.11 &0.72  &0.73  &\textbf{0.82} &0.10 &0.74  & 0.76 &\textbf{0.78} &0.04\\

{\textsc{BERT}} &\textbf{0.78} &0.69 &0.77 &0.09 &\textbf{0.75}  &0.74  &0.73  &0.02 &0.78  &0.69  &\textbf{0.80} &0.11 & \textbf{0.77} &0.74 &0.76 &0.03\\

\textsc{Longformer} &\textbf{0.77} &0.72  &0.73  &0.05 &0.70 &\textbf{0.80} &0.71 &0.10 &0.73 &0.72  &\textbf{0.79} &0.07 & 0.72 & \textbf{0.78} &0.77 &0.06\\

\midrule

\textsc{Avg} & 0.78 & 0.71 & 0.75 & 0.07 & 0.72 & 0.78 & 0.73 & 0.07 & 0.74 & 0.71 & 0.80 & 0.09 & 0.74 & 0.76 & 0.77 & 0.05\\

\bottomrule
\end{tabular}
\caption{\textbf{\em Fairness.} Coverage scores of \texttt{ordered} and \texttt{shuffled} approaches compared to each group of documents ({$\mathcal{D}^w$}, {$\mathcal{D}^a$}, {$\mathcal{D}^h$}) for three extractive summarization models on \emph{DivSumm} dataset. The highest scores are shown in bold.}
\label{tab:extCoverage}
\end{table*}
\endgroup

\begingroup\tabcolsep=3.5pt\def\arraystretch{1}
\begin{table*}[t]
\centering
\small
\begin{tabular}{lccca|ccca|ccca|ccca}
\midrule

\multirow{2}{*}{\textbf{Model}} &  \multicolumn{4}{c}{\textbf{$\mathcal{O}^{w}$}} & \multicolumn{4}{c}{\textbf{$\mathcal{O}^{a}$}}  & \multicolumn{4}{c}{\textbf{$\mathcal{O}^{h}$}}  & \multicolumn{4}{c}{\texttt{shuffled}} \\
\cmidrule{2-17}
& {$\mathcal{D}^w$} & {$\mathcal{D}^a$}  & {$\mathcal{D}^h$} & {$\Delta$Fair} 
& {$\mathcal{D}^w$} & {$\mathcal{D}^a$}  & {$\mathcal{D}^h$} & {$\Delta$Fair} 
& {$\mathcal{D}^w$} & {$\mathcal{D}^a$}  & {$\mathcal{D}^h$} & {$\Delta$Fair} 
& {$\mathcal{D}^w$} & {$\mathcal{D}^a$}  & {$\mathcal{D}^h$} & {$\Delta$Fair}\\
 %\midrule
 %\multicolumn{17}{c}{\textbf{Similarity}}\\
\midrule
\textsc{Textrank} &\textbf{0.57}   &0.55  &0.52  &0.05 &0.51  &\textbf{0.54}   &0.49  &0.05 &\textbf{0.55}  &0.54  &0.50 &0.04  &0.45  & \textbf{0.46} &0.42 &0.05\\

{\textsc{BERT}} &\textbf{0.61}  &0.54  &0.53 &0.07 &0.51  &0.59  &\textbf{0.61}  &0.10 &0.62  &\textbf{0.63}  &0.55 &0.08 & 0.48 &0.50  &\textbf{0.52} &0.03\\

\textsc{LongFormer} &\textbf{0.58}   &0.54  &0.50 &0.08  &0.55 &\textbf{0.56}   &0.55 &0.02 &\textbf{0.54} &\textbf{0.54}  &0.52 &0.02 & 0.45 & 0.44 &\textbf{0.47} &0.03 \\

\midrule

\textsc{Avg} & 0.59 & 0.54 & 0.52 & 0.07 & 0.52 & 0.56 & 0.55 & 0.04 & 0.57 & 0.57 & 0.52 & 0.05 & 0.46 & 0.47 & 0.47 & 0.03\\

\bottomrule
\end{tabular}
\caption{\textbf{\em Fairness.} Semantic  similarity scores of \texttt{ordered} and \texttt{shuffled} approaches compared to each group of documents ({$\mathcal{D}^w$}, {$\mathcal{D}^a$}, {$\mathcal{D}^h$}) across extractive summarization models on \emph{DivSumm} dataset. The highest scores are shown in bold.}
\label{tab:extSimilarity}
\end{table*}
\endgroup

\begingroup\tabcolsep=3.5pt\def\arraystretch{1}
\begin{table*}[t]
\centering
\small
\begin{tabular}{lcccc|cccc|cccc|cccc}
\midrule

\multirow{2}{*}{\textbf{Model}} &  \multicolumn{4}{c}{\textbf{{ROUGE-L}} } & \multicolumn{4}{c}{\textbf{\textsc{BARTScore}} } & \multicolumn{4}{c}{\textbf{\textsc{BERTScore}} } & \multicolumn{4}{c}{\textbf{\textsc{UniEval}} }\\
\cmidrule{2-17}
&$\mathcal{O}^{w}$ & $\mathcal{O}^{a}$  & $\mathcal{O}^{h}$  & \texttt{Sh.} 
&$\mathcal{O}^{w}$ & $\mathcal{O}^{a}$  & $\mathcal{O}^{h}$  & \texttt{Sh.}  
&$\mathcal{O}^{w}$ & $\mathcal{O}^{a}$  & $\mathcal{O}^{h}$  & \texttt{Sh.} 
&$\mathcal{O}^{w}$ & $\mathcal{O}^{a}$  & $\mathcal{O}^{h}$  & \texttt{Sh.}\\
 \midrule

\textsc{Textrank} &\textbf{0.23}  &0.21  &0.22  &\textbf{0.23}  &-4.42  &-4.42  &-4.44  &\textbf{-4.29}  &0.55  &0.54  &0.55  &\textbf{0.56} &0.46 &0.46 &\textbf{0.48} &0.44  \\

{\textsc{BERT}} &\underline{\textbf{0.24}} &\underline{\textbf{0.24}}  &0.23  &0.21  &\underline{\textbf{-4.28}}  &-4.33  &-4.39  &-4.71  &\textbf{0.56}  &\textbf{0.56}  &\textbf{0.56}  &0.55 &0.47 &0.46 &\underline{\bf 0.49} &0.45 \\

\textsc{Longformer} &\textbf{0.22}  &0.21  &\textbf{0.22}  &0.20  &-4.38  &-4.44  &-4.41  &\textbf{-4.35}  &\textbf{0.56}  &0.55  &\textbf{0.56}  &\textbf{0.56} &0.46 &0.46 &\textbf{0.48} &0.45  \\

\midrule

\textsc{Avg} & 0.23 & 0.22 & 0.22 & 0.22 & -4.36 & -4.40 & -4.41 & -4.45 & 0.56 & 0.55 & 0.55 & 0.56 & 0.46 & 0.46 & 0.48 & 0.45\\

\bottomrule
\end{tabular}
\caption{\textbf{\em Quality}. Results of \texttt{ordered} and \texttt{shuffled} approaches across extractive summarization models showing ROUGE-L, BARTScore, BERTScore and UniEval scores on \emph{DivSumm} dataset. The best scores are shown in {\bf bold}, whereas the highest scores per metric are shown as \underline{underlined}.}
\label{tab:extQuality}
\end{table*}
\endgroup

\label{sec:appendix}

\end{document}